\documentclass{article}
\pdfoutput=1
\usepackage[nonatbib,final]{nips_2017}

\usepackage[utf8]{inputenc} % allow utf-8 input
\usepackage[T1]{fontenc}    % use 8-bit T1 fonts
\usepackage{hyperref}       % hyperlinks
\usepackage{url}            % simple URL typesetting
\usepackage{booktabs}       % professional-quality tables
\usepackage{amsfonts}       % blackboard math symbols
\usepackage{nicefrac}       % compact symbols for 1/2, etc.
\usepackage{microtype}      % microtypography
\usepackage{graphicx}
\usepackage{amsmath}
\usepackage{amssymb}
\usepackage[numbers]{natbib}
\usepackage{enumitem}
\usepackage{subcaption}
\usepackage{multirow}
\usepackage{bigstrut}
\usepackage{wrapfig}
\usepackage{adjustbox}

\setdescription{nosep,itemsep=0pt,parsep=0pt,leftmargin=0.5cm}
\setitemize[1]{label=---}
\makeatletter
\renewcommand{\paragraph}{%
  \@startsection{paragraph}{4}%
  {\z@}{0ex \@plus 0ex \@minus .2ex}{-1em}%
  {\normalfont\normalsize\bfseries}%
}
\makeatother

\expandafter\def\expandafter\normalsize\expandafter{%
     \normalsize
     \setlength\abovedisplayskip{0pt}%%one space
     \setlength\belowdisplayskip{0pt}%% one more
     \setlength\abovedisplayshortskip{0pt}%% a third
     \setlength\belowdisplayshortskip{0pt}%% a fourth
}

\makeatletter
  \newcommand\figcaption{\def\@captype{figure}\caption}
  \newcommand\tabcaption{\def\@captype{table}\caption}
\makeatother

\title{Learning to Transfer}

\author{
  Ying Wei$^{\ast}$ \quad\quad\quad\quad\quad\quad\quad\quad Yu Zhang$^{\dagger}$  \quad\quad\quad\quad\quad\quad\quad\quad    Qiang Yang$^{\ddagger}$\\
  Department of Computer Science,
  Hong Kong University of Science and Technology,
  Hong Kong \\
  \texttt{yweiad$^{\ast}$,zhangyu$^{\dagger}$,qyang$^{\ddagger}$@cse.ust.hk} \\
}

\begin{document}
% \nipsfinalcopy is no longer used

\maketitle

\begin{abstract}
Transfer learning borrows knowledge from a source domain to facilitate learning in a target domain.
Two primary issues to be addressed in transfer learning are what and how to transfer.
For a pair of %source and target 
domains, adopting different transfer learning algorithms results in different knowledge transferred between them.
To discover the optimal transfer learning algorithm that maximally improves the learning performance in the target domain, researchers have to exhaustively explore all existing transfer learning algorithms, which is computationally intractable.
As a trade-off, a sub-optimal algorithm is selected, which requires considerable expertise in an ad-hoc way. % and not systematically.
Meanwhile, it is widely accepted in educational psychology that human beings improve transfer learning skills of deciding what to transfer through meta-cognitive reflection on inductive transfer learning practices.
Motivated by this, we propose a novel transfer learning framework known as
 \emph{Learning to Transfer}~(L2T) to automatically determine what and how to transfer %is
are 
the best by leveraging previous transfer learning experiences.
We  establish the L2T framework in two stages: 1) we first learn a reflection function %which 
encrypting transfer learning skills from experiences; and 2) we
infer what and how to transfer for a newly arrived pair of domains by optimizing the %learned 
reflection function. %learned in the first stage.
Extensive experiments %on two image datasets 
demonstrate the L2T's superiority over several state-of-the-art transfer learning algorithms and its effectiveness on discovering more transferable knowledge.

\end{abstract}

\section{Introduction}
\vspace{-0.1in}
Inspired by human beings' capabilities to transfer knowledge across tasks, transfer learning aims to leverage knowledge from a source domain to improve the learning performance or minimize the number of %training
labeled examples required in a target domain.
It is of particular significance when tackling tasks with limited %training
labeled examples.
Transfer learning has proved its wide applicability in, %for example,
such as
 image classification~\cite{Long:Learning}, sentiment classification~\cite{Blitzer:Domain}, dialog systems~\cite{Mo:Personalizing} and urban computing~\cite{Wei:Transfer}.

Three key research issues in transfer learning, pointed by~\citeauthor{Pan:Survey}, are when to transfer, how to transfer, and what to transfer.
Once transfer learning from a source domain is considered to benefit a target domain (when to transfer), an algorithm (how to transfer) should discover the shared knowledge to be transferred across domains (what to transfer).
Different algorithms are likely to discover different parts of transferable knowledge, and thereby lead to uneven transfer learning effectiveness which is evaluated by the performance improvement after transfer learning is conducted.
To achieve the optimal performance improvement for a pair of source and target domains, researchers may try tens to hundreds of transfer learning algorithms covering instance~\cite{Dai:Boosting}, parameter~\cite{Tommasi:Learning}, and feature~\cite{Pan:TCA} based algorithms.
%The current transfer learning algorithms determine have determined what and how to trans- fer based solely on the pair of source and target domains at hand.
%The variance in the knowledge selected for different algorithms to transfer, in the case, could be very high.
%Researchers must explore tens or hundreds of algorithms to achieve an optimal transfer,
This brute-force exploration is computationally expensive and 
%even 
practically impossible.
As a tradeoff, a sub-optimal improvement is usually obtained from a heuristically selected algorithm, which 
unfortunately 
requires considerable expertise in an ad-hoc and unsystematic manner.

Exploring different algorithms is not the only way to better determine what to transfer and thereby improve transfer learning effectiveness.
Previous transfer learning experiences do also help, which has been widely accepted in educational psychology~\cite{Belmont:Secure,Luria:Cognitive}.
Human beings sharpen transfer learning skills for deciding what to transfer by conducting meta-cognitive reflection on thorough and diverse transfer learning experiences.
For example, children who are good at playing chess may transfer mathematical skills, reading skills, visuospatial skills, and decision making skills learned from chess to solve arithmetic problems, to comprehend reading materials, to solve pattern matching puzzles, and to play basketball, respectively.
At a later age, it will be easier for them to decide to transfer mathematical and decision making skills learned from chess, rather than reading and visuospatial skills, to market investment.
%In e-commerce, the genre and topics of the books that users have previously chosen can be transferred to understand the users’ interest in movies. Instead, we transfer the reviews and comments of those books to predict the users’ preferences for kitchen products.
Unfortunately, all existing transfer learning algorithms transfer from scratch and ignore previous transfer learning experiences.

Motivated by this, we propose a novel transfer learning framework called Learning to Transfer~(L2T).
The key idea of the L2T framework is to enhance transfer learning effectiveness by leveraging previous transfer learning experiences to automatically determine what and how to transfer are the best for a pair of source and target domains of interest.
To achieve the goal, we establish the L2T framework in two stages.
During the first stage, we encode each transfer learning experience into three components including a pair of source and target domains, transferred knowledge parameterized as shared latent feature factors (i.e., what to transfer between them), and performance improvement.
We learn from all experiences a reflection function which maps a pair of domains and the transferred knowledge between them to the corresponding performance improvement.
The reflection function, therefore, is believed to encrypt the transfer learning skills for deciding what and how to transfer.
In the second stage, the knowledge to be transferred %for
between a newly arrived pair of %source and target 
domains is optimized so
 that the value of the learned reflection function, matching to the performance improvement, is maximized.

The contribution %s
of this paper 
%are two-fold. Firstly 
is %lies in 
that 
we propose a novel transfer learning framework which opens a new door to improve %the 
transfer learning effectiveness
%first 
by taking advantages of previous transfer learning experiences.
%into consideration
%. Consequently, the L2T framework 
The L2T %framework 
can 
%automatically and systematically 
discover more transferable knowledge across domains %and greatly improves the transfer learning effectiveness
in a systematic and automatic fashion without requiring considerable expertise,  
%Secondly we demonstrate that with 
which is also evidenced in 
comprehensive empirical %evidence
studies 
showing the 
%demonstrates that
%the
L2T's superiority over %framework 
%outperforms 
state-of-the-art transfer learning algorithms. % without meticulous algorithm selection.

%To deal with L2T, we will be facing several critical research challenges. First, given a pair of source and target domains, how do we design an automatic L2T framework? Second, how do we implement the L2T framework when considering a single level of latent feature factors as “what and how to transfer?” Third, how do we extend the form of the transferred knowledge to give it more flexibility, i.e., hierarchical latent factors, to satisfy more complex transfer experiences?

%As a new transfer learning framework, our proposed L2T framework will have significant impacts. For any application traditional transfer learning has explored, including image clas- sification [22] and WiFi localization [27], L2T can be applied to foster transfer performance. Moreover, our L2T framework will automate the transfer learning process and eliminate metic- ulous algorithm selection, thereby reaching out to other communities of researchers.

\setlength{\textfloatsep}{1pt plus 1.0pt minus 2.0pt}
\section{Related Work}
\vspace{-0.1in}
%In this section, we will review the %three
%two
%related strands of work, including transfer learning %, %multi-task learning,
%and lifelong learning.
%, and demonstrate the differences between L2T with them in Figure~\ref{fig:related_work}.

%\begin{figure}[h]
%\centering
%\includegraphics[scale = 0.44]{fig/related_work}
%\caption{Illustration of the differences between our work and  the other three lines of work.}
%\label{fig:related_work}
%\end{figure}
\paragraph{Transfer Learning}
%Transfer learning, accounting for human beings' ability to borrow knowledge from domains previously learnt to those newly arrived, has drawn increased attention recently.
To successfully conduct transfer learning, one of the most critical research issues identified in~\cite{Pan:Survey} is to decide what and how to transfer.
Parameters~\cite{Tommasi:Learning,Yang:Adapting},
instances~\cite{Dai:Boosting},
or latent feature factors~\cite{Pan:TCA}
can be transferred between domains.
A few works~\cite{Tommasi:Learning,Yang:Adapting}  transfer \emph{parameters} from source domains to a target domain as regularizers of SVM-based models. % with few labeled training examples.
In \cite{Dai:Boosting}, a basic learner in a target domain is boosted by borrowing the most useful \emph{instances} from a source domain.
Different techniques capable of extracting transferable \emph{latent feature factors} between domains have been investigated extensively.
These techniques include manually selected pivot features~\cite{Blitzer:Domain}, dimension reduction~\cite{Baktashmotlagh:Unsupervised,Baktashmotlagh:Domain,Pan:TCA},   collective matrix factorization~\cite{Long:Graph}, dictionary learning and sparse coding~\cite{Raina:Self,Zhang:LSDT}, manifold learning~\cite{Gong:Geodesic,Gopalan:Domain}, and deep learning~\cite{Long:Learning,Tzeng:Simultaneous,Yosinski:Transferable}.
%The pioneering work transferred manually selected pivot features which are domain invariant.
%\citeauthor{Pan:TCA} developed the transfer component analysis to reduce the distance between domains in the latent space spanned by transfer components.
Unlike L2T,
%our learning to transfer framework,
all existing studies in transfer learning
transfer from scratch, i.e.,
%aim at
%improving the learning performance for a target task %(e.g., Task 2 in Figure~\ref{fig:related_work})
%by
only considering the source domain and target domain of interest
but ignoring previous transfer learning experiences. %(e.g., Task 1).
%Instead, the goal of our framework is to improve the transfer effectiveness for a pair of source and target tasks (e.g., Task 2N+1 and Task 2N+2) by learning meta-knowledge from previous transfer learning experiences.
Since any existing transfer learning algorithm mentioned above could be %adopted
applied
 in a transfer learning experience,
L2T can even collect all algorithms' wisdom together.%\\
%\textbf{Multi-task Learning}
%Multi-task learning~\cite{Caruana:Multitask}  trains multiple related tasks simultaneously and learns shared knowledge among tasks, so that the generalization capabilities of all tasks reinforce each other.
%In almost all multi-task learning work~\cite{Argyriou:Multi}, the shared knowledge are parameters.
%%Modelling task relationships among tasks~\cite{Zhang:Convex} is key to improve learning common parameters.
%However, multi-task learning assumes that the training and test data follow the same distribution, as Figure~\ref{fig:related_work} shows, which is different from transfer learning we focus on. \\
\paragraph{Lifelong Learning}
%By 
Assuming a new learning task %(e.g., the (N+1)-th task) % N+1 in Figure~\ref{fig:related_work})
to lie in the same environment as training tasks, %(e.g. N tasks),
learning to learn~\cite{Thrun:Learning} or meta-learning~\cite{Maurer:Algorithmic} transfers knowledge shared among the training tasks to the new task. \citeauthor{Ruvolo:Ella} considered lifelong learning as an online meta-learning.
%in which knowledge learnt from previous tasks is transferred to a newly arrived task over time.
Though L2T and lifelong learning both aim to continuously improve a learning system by leveraging histories, L2T differs from them in that each task we consider is a transfer learning task rather than
 a traditional learning task.
Therefore, we learn transfer learning skills instead of task-sharing knowledge.

\section{Learning to Transfer}
\vspace{-0.1in}
We begin by first introducing the proposed L2T framework.
Then we detail the two steps involved in the framework, i.e., learning transfer learning skills from previous transfer learning experiences and applying those skills to infer what and how to transfer for a future pair of source and target domains.

%In this section, we first introduce the proposed L2T framework, followed by the two steps involved in the framework, i.e., learning transfer learning skills from previous transfer learning experiences, and applying those skills to infer what and how to transfer for a future pair of source and target domains.
\subsection{The L2T Framework}
\label{sec:framework}
\begin{wrapfigure}{r}{0.7\textwidth}
\vspace{-0.2in}
\centering
\includegraphics[scale = 0.34]{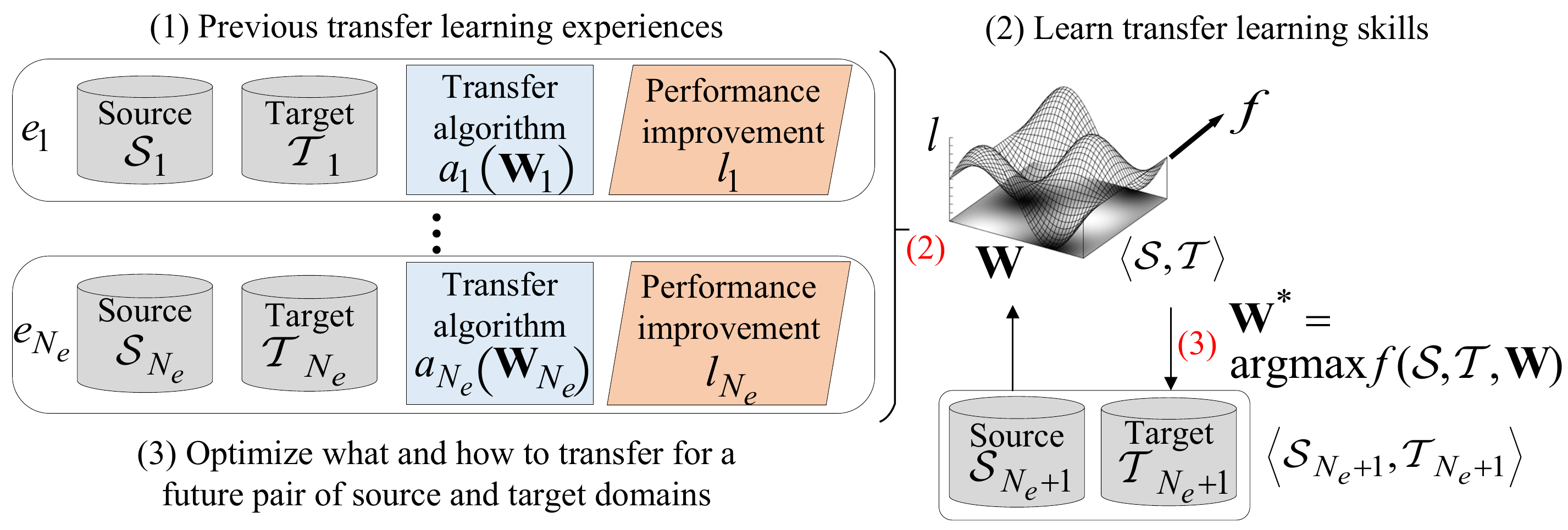}
\caption{Illustration of the L2T framework.}
\label{fig:framework}
\vspace{-0.13in}
%\end{figure}
\end{wrapfigure}

A L2T agent previously conducted transfer learning several times, and kept a record of $N_e$ transfer learning experiences (see step (1) in Figure~\ref{fig:framework}).
We define each transfer learning experience as $E_{e}=(\mathcal{S}_e,\mathcal{T}_e,a_e,l_e)$ in which $\mathcal{S}_e=\{\mathbf{X}^s_e,\mathbf{y}^s_e\}$  and $\mathcal{T}_e=\{\mathbf{X}^t_e,\mathbf{y}^t_e\}$ denote a source domain and a target domain, respectively.
$\mathbf{X}^*_e\in\mathbb{R}^{n_e^*\times m}$ represents the feature matrix if either domain has $n^*_e$ examples in a $m$-dimensional feature space $\mathcal{X}_e^*$, where the superscript $*$ can be either $s$ or $t$ to denote %the
a source or target domain.
$\mathbf{y}^*_e\in\mathcal{Y}^*_e$ denotes the vector of labels with the length being $n_{le}^{*}$.
The number of target labeled examples is much smaller than that of source labeled examples, i.e., $n_{le}^{t}\ll n_{le}^{s}$.
We focus on the setting $\mathcal{X}_e^s = \mathcal{X}_e^t$ and $\mathcal{Y}_e^s\neq \mathcal{Y}_e^t$
for each pair of domains.
%for each transfer learning experience.
$a_e\in \mathcal{A}=\{a_1, \cdots, a_{N_a}\}$ denotes a transfer learning algorithm randomly selected from the set $\mathcal{A}$ containing $N_a$ base algorithms.
Suppose that what to transfer inferred by the algorithm $a_e$ can be parameterized as $\mathbf{W}_e$.
%transferred latent feature factors, i.e., what to transfer, inferred by the transfer learning algorithm run in the $e$-th experience.
Finally, each transfer learning experience is labeled with the performance improvement ratio $l_e=p^{st}_e/p^t_e$ where $p^{t}_e$ is the learning performance (e.g., classification accuracy) in $\mathcal{T}_e$ without transfer on a test dataset and $p^{st}_e$ is the learning performance in $\mathcal{T}_e$ after transferring knowledge from $\mathcal{S}_e$ on the same test dataset.

%\begin{figure}[!h]
With all previous transfer learning experiences as the input, the L2T agent aims to learn a function $f$ such that $f(\mathcal{S}_e, \mathcal{T}_e, \mathbf{W}_e)$ approximates $l_e$ as shown in step (2) of Figure~\ref{fig:framework}.
We call $f$ a \emph{reflection} function which encrypts the meta-cognitive transfer learning skills - what and how to transfer can maximize the improvement ratio given a pair of source and target domains.
Whenever a new pair of domains $\langle\mathcal{S}_{N_e+1},\mathcal{T}_{N_e+1}\rangle$ arrives,
the L2T agent can identify the optimal knowledge %latent feature factors
to be transferred for this new pair, which is contained in $\mathbf{{W}}^*_{N_e+1}$,  by maximizing $f$ (i.e., step (3) in Figure~\ref{fig:framework}).

\subsection{Parameterizing What to Transfer}
\label{sec:para}
%Why parametrize?
%As reviewed in the section of related work, what to transfer between domains can be categorized into parameters, instances, and latent feature factors.
Adopting different algorithms
%within each category
for one pair of source and target domains brings different what to transfer between them. %and thereby generates different experiences.
To learn the reflection function with
%mapping a pair of domains and
what to transfer as one input,
%to the improvement ratio,
we have to uniformly parameterize ``what to transfer'' for all algorithms in the base algorithm set $\mathcal{A}$.
In this work, we consider  $\mathcal{A}$ to contain algorithms transferring only single-level latent feature factors, because
%First, %the reason why we exclude
other existing parameter-based and instance-based algorithms
%is that existing work in these two categories
cannot address the transfer learning setting we focus on
%- a pair of domains share the feature space but have disparate label spaces
(i.e., $\mathcal{X}^e_s=\mathcal{X}^e_t$ and $\mathcal{Y}^e_s\neq\mathcal{Y}^e_t$).
Though limited parameter-based algorithms~\cite{Tommasi:Learning,Yang:Adapting} can %do
transfer across domains in heterogeneous label spaces, they can only handle binary classification problems.
Deep neural network based algorithms~\cite{Long:Learning,Tzeng:Simultaneous,Yosinski:Transferable} transferring latent feature factors in a hierarchy are left for our future research.
%How parametrize?
%By restricting the base transfer learning algorithm set $\mathcal{A}$ to contain only algorithms transferring single level of latent feature factors, we parametrize what to transfer
%of different experiences
%with the latent feature factor matrix $\mathbf{W}$.
%Though limited paramter based algorithms~\cite{Yang:Adapting,Tommasi:Learning} can transfer across domains in heterogeneous label spaces, they can only deal with binary classification problems.
%In the next, we will detail how ``what to transfer'' inferred from all base transfer learning algorithms in the set $\mathcal{A}$  can be uniformly parametrized by $\mathbf{W}$.
As a consequence, what to transfer can be parameterized with a latent feature factor matrix $\mathbf{W}$, which we will elaborate in the following.

%\begin{figure}[b]
%  \begin{minipage}[t]{0.45\textwidth}
%    \centering
%    \includegraphics[scale=0.4]{fig/feature_transfer_algorithms}
%    \figcaption{Summary of two types of latent feature factor based transfer learning algorithms.}
%	\label{fig:feature_transfer}
%  \end{minipage}%
%  \hspace{0.1in}
%  \begin{minipage}[t]{0.45\textwidth}
%    \centering
%    \includegraphics[scale=0.24]{fig/phi}
%    \figcaption{A Pictorial illustration of the nonlinear mapping function $\phi$ to the RKHS.}
%	\label{fig:phi}
%  \end{minipage}%
%\end{figure}

%\subsubsection{Base transfer learning algorithms}
%The majority of transductive transfer learning algorithms fall into this category.
Latent feature factor based algorithms
for the transfer learning setting mentioned above
%latent feature factor based algorithms
aim to
%The key intuition for latent feature factor based algorithms is to
learn domain-invariant
feature factors
%feature representation
across domains.
Consider classifying dog pictures as a source domain and cat pictures as a target domain.
The domain-invariant feature factors could include eyes, mouth, tails, etc.
%With both source and target examples re-characterized with this representation, a model trained on source examples can apply to target examples either.
Re-characterized with the latent feature factors which are more descriptive, a target domain is expected to achieve better performance with less labeled examples.
What to transfer, in this case, is these shared feature factors across domains.
% this place needs clarification of deep learning based methods
According to different ways of extracting the domain-invariant feature factors,
existing latent feature factor based algorithms can be categorized into two groups, i.e.,
%according to the structure of the shared feature representation:
common latent space based algorithms and manifold ensemble based algorithms. % and subspace alignment.
%We characterize the two types in Figure~\ref{fig:feature_transfer}.

%\vspace{-0.1in}
%\begin{enumerate}[topsep=0in,leftmargin=*,itemindent=0.18in]
\paragraph{Common latent space based algorithms}
This line of algorithms, including but not limited to TCA~\cite{Pan:TCA}, LSDT~\cite{Zhang:LSDT}, and DIP~\cite{Baktashmotlagh:Unsupervised},
%, and TSC~\cite{Long:Transfer},
assumes
that the domain-invariant feature factors lie in
%the existence of
a single shared latent space.
%where the domain-invariant feature factors lie.
We denote by $\varphi$ the function mapping the original feature representation into the latent space. %defined
%, the cosine similarity between a pair of examples $\mathbf{x}_i$ and $\mathbf{x}_j$ in the space is defined by $s_{\varphi}=\frac{\varphi(\mathbf{x}_i)^T\varphi(\mathbf{x}_j)}{\Vert\varphi(\mathbf{x}_i)^T\varphi(\mathbf{x}_i)\Vert\cdot\Vert\varphi(\mathbf{x}_j)^T\varphi(\mathbf{x}_j)\Vert}$.
%producing different latent spaces would incur different similarity matrices.
If $\varphi$ is linear, it can be represented %by
as
 an embedding matrix $\mathbf{W}\in\mathbb{R}^{m\times u}$ where $u$ is the dimensionality of the latent space.
%, $s_{\varphi}=s_{\mathbf{W}}=\frac{\mathbf{x}_i^T\mathbf{W}^T\mathbf{W}\mathbf{x}_j^T}{\sqrt{\mathbf{x}_i^T\mathbf{W}^T\mathbf{W}\mathbf{x}_i^T}\sqrt{\mathbf{x}_j^T\mathbf{W}^T\mathbf{W}\mathbf{x}_j^T}}$.
%By introducing $\mathbf{G}=\mathbf{W}^T\mathbf{W}\in\mathbb{R}^{m\times m}$ widely known as the metric matrix in similarity metric learning~\cite{Cao2013,Chechik2010}, we rewrite the similarity as $s_{\mathbf{G}}=\frac{\mathbf{x}_i^T\mathbf{G}\mathbf{x}_j}{\sqrt{\mathbf{x}_i^T\mathbf{G}\mathbf{x}_i}\sqrt{\mathbf{x}_j^T\mathbf{G}\mathbf{x}_j}}$.
%Different $\mathbf{W}$ will produce different shared latent subspaces.
Therefore, we can parameterize the latent space, what to transfer we focus on, with $\mathbf{W}$ which describes the $u$ latent feature factors. %using the metric matrix $\mathbf{G}$.
Otherwise, if $\varphi$ is nonlinear, the latent space can still be parameterized with $\mathbf{W}$.
Although a nonlinear  $\varphi$ is not explicitly specified in most cases such as %kernel based TCA and
LSDT using sparse coding, transformed target instances %represented 
in the latent space $\mathbf{Z}^t_e=\varphi(\mathbf{X}^t_e)\in\mathbb{R}^{n^t_e\times u}$ are always available.
Consequently, we can infer the similarity metric matrix~\cite{Cao2013} in the latent space, i.e., $\mathbf{G}\in\mathbb{R}^{m\times m}$, according to $\mathbf{X}^t_e\mathbf{G}(\mathbf{X}^t_e)^T=\mathbf{Z}^t_e(\mathbf{Z}^t_e)^T$. %where $\mathbf{G}$ is the similarity matrix in metric learning~\cite{Cao2013},
%we compute %$\mathbf{G}$,
% in the latent space, i.e.,
The solution $\mathbf{G}=(\mathbf{X}^t_e)^{\dagger}\mathbf{Z}^t_e(\mathbf{Z}^t_e)^T[(\mathbf{X}^t_e)^T]^{\dagger}$
where $(\mathbf{X}^t_e)^{\dagger}$ is the psudo-inverse of $\mathbf{X}^t_e$.
By performing LDL decomposition on $\mathbf{G}=\mathbf{L}\mathbf{D}\mathbf{L}^T$, we infer the latent feature factor matrix $\mathbf{W} = \mathbf{L}\mathbf{D}^{1/2}$.
%is inferred. this place needs to be confirmed.
\paragraph{Manifold ensemble based algorithms}
Initiated by~\citeauthor{Gopalan:Domain}, manifold ensemble algorithms consider that a source and a target domain share multiple subspaces (of the same dimension) as points on the Grassmann manifold between them.
The %domain-invariant
representation of a target domain on $u$ domain-invariant latent factors turns to
%$\mathbf{X}_e^t$
%is
$\mathbf{Z}_e^{t(n_u)}=[\varphi_1(\mathbf{X}_e^t),  \cdots, \varphi_{n_u}(\mathbf{X}_e^t)]\in\mathbb{R}^{n^t_e\times n_u u}$,
%provided that
%\varphi_2(\mathbf{X}_e^t),
if $n_u$ subspaces on the manifold are sampled.
%Even
When all continuous subspaces on the manifold are considered, i.e., $n_u\rightarrow\infty$, ~\citeauthor{Gong:Geodesic} proved that $\mathbf{Z}^{t(\infty)}_e(\mathbf{Z}^{t(\infty)}_e)^T = \mathbf{X}^t_e\mathbf{G}(\mathbf{X}^t_e)^T$ where $\mathbf{G}$ is the similarity metric matrix. %abovementioned. %in similarity metric learning~\cite{Cao2013}.
%$\frac{(\mathbf{z}^{\infty}_i)^T\mathbf{z}^{\infty}_j}{\Vert\mathbf{z}^{\infty}_i\Vert\cdot\Vert\mathbf{z}^{\infty}_j\Vert}=\frac{\mathbf{x}_i^T\mathbf{G}\mathbf{x}_j}{\sqrt{\mathbf{x}_i^T\mathbf{G}\mathbf{x}_i}\sqrt{\mathbf{x}_j^T\mathbf{G}\mathbf{x}_j}}$ where $\mathbf{G}$ can be computed in a closed form.
For computational details of $\mathbf{G}$, please refer to~\cite{Gong:Geodesic}.
%Different manifolds will have different $\mathbf{G}$.
$\mathbf{W}=\mathbf{L}\mathbf{D}^{1/2}$, therefore, is also qualified to represent the latent feature factors distributed in a series of subspaces on a manifold. %to be transferred.

\subsection{Learning from Experiences}
The goal here is to learn a reflection function $f$ such that $f(\mathcal{S}_e, \mathcal{T}_e, \mathbf{W}_e)$ can approximate $l_e$ for all experiences $e\in\{1,\cdots,N_e\}$.
The %performance
improvement ratio $l_e$ is closely related to two aspects: 1) the difference between a source and a target domain in the shared latent space, and 2) the discriminative ability of a target domain in the %shared
latent space.
The smaller difference guarantees
more overlap between domains
%a source and a target domain
in the %shared
latent space, which signifies more transferable latent feature factors and higher %performance
improvement ratios as a result.
%Higher improvement ratio $l_e$ indicates that the latent factors $\mathbf{W}_e$ are more transferable.
The more discriminative %cap
ability of a target domain in the latent space is also vital to improve %learning
performances.
%With more transferable latent factors $\mathbf{W}_e$, the distance between new representations of a source and a target domain in the shared latent space, i.e., $\mathbf{X}^s_e\mathbf{W}_e$ and $\mathbf{X}^t_e\mathbf{W}_e$, will get smaller.
Therefore, we build $f$ to take %the
both aspects into consideration.

\paragraph{The difference between a source and a target domain}
We follow~\cite{Baktashmotlagh:Unsupervised,Pan:TCA} and adopt the maximum mean discrepancy (MMD)~\cite{Gretton:Optimal} to measure the difference between a source and a target domain.
By mapping two domains into the reproducing kernel Hilbert space (RKHS), MMD
empirically
evaluates the distance between the means of instances from a source domain and a target domain:
\begin{small}
\begin{align}
\hat{d}_e^2(\mathbf{X}^s_e\mathbf{W}_e,\mathbf{X}^t_e\mathbf{W}_e) =&\bigg\Vert\frac{1}{n^s_e}\sum_{i=1}^{n^s_e}\phi(\mathbf{x}^{s}_{ei}\mathbf{W}_e)-\frac{1}{n^t_e}\sum_{j=1}^{n^t_e}\phi(\mathbf{x}^{t}_{ej}\mathbf{W}_e)\bigg\Vert_{\mathcal{H}}^2
 =\frac{1}{(n^s_e)^2}\sum_{i,i'=1}^{n^s_e}\mathcal{K}(\mathbf{x}^{s}_{ei}\mathbf{W}_e, \mathbf{x}^{s}_{ei'}\mathbf{W}_e)\nonumber \\
& +\frac{1}{(n^t_e)^2}\sum_{j,j'=1}^{n^t_e}\mathcal{K}(\mathbf{x}^{t}_{ej}\mathbf{W}_e,\mathbf{x}^{t}_{ej'}\mathbf{W}_e)
-\frac{2}{n^s_e n^t_e}\sum_{i,j=1}^{n^s_e,n^t_e}\mathcal{K}(\mathbf{x}^{s}_{ei}\mathbf{W}_e,\mathbf{x}^{t}_{ej}\mathbf{W}_e),
\label{eqn:mmd}
\end{align}
\end{small}\noindent
where $\phi$ maps from the $u$-dimensional %shared
latent space to the %high-dimensional
RKHS $\mathcal{H}$ and $\mathcal{K}(\cdot,\cdot)=\langle\phi(\cdot),\phi(\cdot)\rangle$ is the kernel function.
%Alternatively in a matrix form,
%\begin{equation}
%d_e^2(\mathbf{X}^s_e,\mathbf{X}^t_e)=tr(\mathbf{K}_{\mathbf{W}_e}\mathbf{M}_e),
%\end{equation}
%where $\mathbf{K}_{\mathbf{W}_e}= \begin{bmatrix}
%    \mathcal{K}(\mathbf{X}_e^s\mathbf{W}_e, \mathbf{X}_e^s\mathbf{W}_e)  & \mathcal{K}(\mathbf{X}_e^s\mathbf{W}_e, \mathbf{X}_e^t\mathbf{W}_e) \\
%   \mathcal{K}(\mathbf{X}_e^t\mathbf{W}_e, \mathbf{X}_e^s\mathbf{W}_e) & \mathcal{K}(\mathbf{X}_e^t\mathbf{W}_e, \mathbf{X}_e^t\mathbf{W}_e)
%  \end{bmatrix}\in\mathbb{R}^{(n^s_e+n^t_e)\times (n^s_e+n^t_e)}$ and $\mathcal{K}(\cdot,\cdot)=\langle\phi(\cdot),\phi(\cdot)\rangle$ is the kernel function. %$\mathbf{M}$ is computed as
%$\mathbf{M}_e=\begin{bmatrix}
%    \frac{1}{(n^s_e)^2}\mathbf{J}_{n^s_e \times n^s_e} &  -\frac{1}{n^s_e n^t_e}\mathbf{J}_{n^s_e \times n^t_e} \\ -\frac{1}{n^s_e n^t_e}\mathbf{J}_{n^t_e \times n^s_e}
%    & \frac{1}{(n^t_e)^2}\mathbf{J}_{n^t_e \times n^t_e}
%  \end{bmatrix}
%$ in which $\mathbf{J}$ represents an integer matrix consisting of all $1$s.
Using different nonlinear mappings $\phi$ (or equivalently different kernels $\mathcal{K}$) leads to different MMD distances and thereby
different values of $f$.
Therefore, learning the reflection function $f$ is equivalent to finding out the optimal $\phi$ (or $\mathcal{K}$)  such that the MMD distance can well characterize the improvement ratio $l_e$ for all pairs of domains. %as Figure~\ref{fig:phi} shows.
Considering the difficulty of directly defining and learning $\phi$, we learn the optimal $\mathcal{K}$ alternatively.
% leads to different kernels $\mathcal{K}$ and hence different MMD distances, even for one pair of source and target domains.
%In this case, $\mathcal{K}$ can be regarded as a distance metric parameter.
%Learning the reflection function $f$ is equivalent to finding out the optimal $\mathcal{K}$ such that the MMD distance between domains of each experience approximates the improvement ratio well.
Inspired by multiple kernel MMD~\cite{Gretton:Optimal}, we parameterize $\mathcal{K}$ as a combination of $N_{k}$ candidate PSD kernels, i.e., $\mathcal{K}=\sum_{k=1}^{N_k}\beta_k\mathcal{K}_k$ where $\beta_k\geq 0, \forall k$,
and learn the combination coefficients $\boldsymbol{\beta}=[\beta_1,\cdots,\beta_{N_{k}}]$ instead.
%The probabilistic simplex constraint imposed on $\boldsymbol{\beta}$ guarantees $\mathcal{K}$ to be characteristic~\cite{Gretton:Optimal}.
As a result, the MMD %distance 
can be rewritten as $\hat{d}_{e}^2(\mathbf{X}^s_e\mathbf{W}_e,\mathbf{X}^t_e\mathbf{W}_e) = \sum_{k=1}^{N_k}\beta_k \hat{d}_{e(k)}^2(\mathbf{X}^s_e\mathbf{W}_e,\mathbf{X}^t_e\mathbf{W}_e)=\boldsymbol{\beta}^T\mathbf{\hat{d}}_e$, where $\mathbf{\hat{d}}_e=[\hat{d}_{e(1)}^2, \cdots, \hat{d}_{e(N_k)}^2]$ with $\hat{d}_{e(k)}^2$ computed using the $k$-th kernel $\mathcal{K}_k$. %(\mathbf{X}^s_e\mathbf{W}_e,\mathbf{X}^t_e\mathbf{W}_e)
In this paper, we consider %multiple
$N_k$
RBF kernels \begin{small}$\mathcal{K}_k(\mathbf{a},\mathbf{b})=\text{exp}(-\Vert \mathbf{a}-\mathbf{b}\Vert^2/\delta_k)$\end{small} with different bandwidths by varying values of $\delta_k$.

Unfortunately, the MMD only measuring the distance between the means %across
of domains is insufficient to measure the difference between two domains.
The distance variance among all pairs of instances across domains is also required to fully characterize the difference,
% between a source and a target domain,
because  a pair of %source and target
domains with a small MMD but high variance still have a small overlap.
%Inspired by%the two-sample tests
According to~\cite{Gretton:Optimal}, Equation~(\ref{eqn:mmd}) is actually the empirical estimation of
%, we represent the MMD in Equation~\ref{eqn:mmd} as
\begin{small}
$d_{e}^2(\mathbf{X}^s_e\mathbf{W}_e,\mathbf{X}^t_e\mathbf{W}_e)%=\mathbf{E}_{\mathbf{x}^s_{e}\mathbf{x}^{s'}_{e}}\mathcal{K}(\mathbf{x}^s_{ei}\mathbf{W}_{e},\mathbf{x}^s_{ei'}\mathbf{W}_e) + \mathbf{E}_{\mathbf{x}^t_{ej}\mathbf{x}^t_{ej'}}\mathcal{K}(\mathbf{x}^t_{ej}\mathbf{W}_{e},\mathbf{x}^t_{ej'}\mathbf{W}_e)- 2\mathbf{E}_{\mathbf{x}^s_{ei}\mathbf{x}^t_{ej}} \mathcal{K}(\mathbf{x}^s_{ei}\mathbf{W}_{e},\mathbf{x}^t_{ej}\mathbf{W}_e)
=\mathbf{E}_{\mathbf{x}^s_{e}\mathbf{x}^{s\prime}_{e}\mathbf{x}^t_{e}\mathbf{x}^{t\prime}_{e}}h(\mathbf{x}^s_{e},\mathbf{x}^{s\prime}_{e},\mathbf{x}^t_{e},\mathbf{x}^{t\prime}_{e})$
\end{small}
where
\begin{small}
$h(\mathbf{x}^s_{e},\mathbf{x}^{s\prime}_{e},\mathbf{x}^t_{e},\mathbf{x}^{t\prime}_{e})=\mathcal{K}(\mathbf{x}^s_{e}\mathbf{W}_{e},\mathbf{x}^{s\prime}_{e}\mathbf{W}_e)+\mathcal{K}(\mathbf{x}^t_{e}\mathbf{W}_{e},\mathbf{x}^{t\prime}_{e}\mathbf{W}_e)-\mathcal{K}(\mathbf{x}^s_{e}\mathbf{W}_{e},\mathbf{x}^{t\prime}_{e}\mathbf{W}_e)-\mathcal{K}(\mathbf{x}^{s\prime}_{e}\mathbf{W}_{e},\mathbf{x}^t_{e}\mathbf{W}_e)$.
\end{small}
%-\mathbf{E}_{\mathbf{x}^t_{ei'}\mathbf{x}^s_{ej}} \mathcal{K}(\mathbf{x}^t_{ei'}\mathbf{W}_{e},\mathbf{x}^s_{ej}\mathbf{W}_e)
Consequently, the distance variance, $\sigma_e^2$, equals
%\begin{scriptsize}
\begin{equation*}
\begin{adjustbox}{max width=\textwidth}
$\sigma_e^2(\mathbf{X}^s_e\mathbf{W}_e,\mathbf{X}^t_e\mathbf{W}_e)=\mathbf{E}_{\mathbf{x}^s_{e}\mathbf{x}^{s\prime}_{e}\mathbf{x}^t_{e}\mathbf{x}^{t\prime}_{e}}\left[\left(h(\mathbf{x}^s_{e},\mathbf{x}^{s\prime}_{e},\mathbf{x}^t_{e},\mathbf{x}^{t\prime}_{e})-\mathbf{E}_{\mathbf{x}^s_{e}\mathbf{x}^{s\prime}_{e}\mathbf{x}^t_{e}\mathbf{x}^{t\prime}_{e}}h(\mathbf{x}^s_{e},\mathbf{x}^{s\prime}_{e},\mathbf{x}^t_{e},\mathbf{x}^{t\prime}_{e})\right)^2\right].$
\end{adjustbox}
\end{equation*}
%\end{scriptsize}\noindent
To be consistent with the MMD %distance
 characterized with $N_k$ candidate kernels, we can also rewrite \begin{small}
$\sigma_e^2=\boldsymbol{\beta}^T\mathbf{Q}_e\boldsymbol{\beta}$\end{small} where
\begin{small}
$\mathbf{Q}_e=\text{cov}(h)=
\begin{bmatrix}
   \sigma_{e(1,1)} & \cdots & \sigma_{e(1 ,N_{k})} \\
   \cdots & \cdots & \cdots \\
   \sigma_{e(N_k, 1)} & \cdots & \sigma_{e(N_k ,N_k)}
  \end{bmatrix}$.
\end{small}
Each \mbox{element} \begin{small}$\sigma_{e(k_1,k_2)}=
\text{cov}(h_{k_1}, h_{k_2})=\mathbf{E}\left[(h_{k_1}-\mathbf{E}h_{k_1})(h_{k_2}-\mathbf{E}h_{k_2})\right]$\end{small}.
Note that
\begin{small}
$\mathbf{E}h_{k_1}$\end{small} is shorthand for
\begin{small}
$\mathbf{E}_{\mathbf{x}^s_{e}\mathbf{x}^{s\prime}_{e}\mathbf{x}^t_{e}\mathbf{x}^{t\prime}_{e}}h_{k1}(\mathbf{x}^s_{e},\mathbf{x}^{s\prime}_{e},\mathbf{x}^t_{e},\mathbf{x}^{t\prime}_{e})$
\end{small}
where \begin{small}
$h_{k_1}$
\end{small}
is calculated using the $k_1$-th kernel. Due to unknown data distributions of either source or target domains, we empirically estimate $\mathbf{Q}_e$ to be $\mathbf{\hat{Q}}_e$
which is
 detailed in the supplementary material due to page limit.

\paragraph{The discriminative ability of a target domain}
Considering the %serious
 lack of labeled data in a target domain, we resort to unlabeled data to evaluate the discriminative ability instead of using labelled data %to directly measure
as usual.
The principles of the unlabeled discriminant criterion are two-fold: 1) similar instances %of a target domain %in the original feature space
should also be neighbours after being embedded into the latent space; and 2) dissimilar instances should be far away.
We adopt the unlabeled discriminant criterion proposed in~\cite{Yang:Globally},
\begin{small}
\begin{equation*}
\tau_e = \text{tr}(\mathbf{W}_e^T\mathbf{S}_e^N\mathbf{W}_e)/\text{tr}(\mathbf{W}_e^T\mathbf{S}_e^L\mathbf{W}_e),
\end{equation*}
\end{small}\noindent
where \begin{small}$\mathbf{S}_e^L=\sum_{j,j'=1}^{n^t_e}\frac{H_{jj'}}{(n^t_e)^2}(\mathbf{x}^t_{ej}-\mathbf{x}^t_{ej'})(\mathbf{x}^t_{ej}-\mathbf{x}^t_{ej'})^T$\end{small}
%\sum_{j'=1}^{n^t_e}
is the local scatter covariance matrix with the neighbour information
$H_{jj'}$ defined as
\begin{small}
$
H_{jj'}=\begin{cases}
    \mathcal{K}(\mathbf{x}_{ej}^t,\mathbf{x}_{ej'}^t), & \text{if} \quad \mathbf{x}^t_{ej} \in\mathcal{N}_r(\mathbf{x}^t_{ej'}) \quad \text{and}\quad \mathbf{x}^t_{ej'} \in\mathcal{N}_r(\mathbf{x}^t_{ej})\\
    0, & \text{otherwise}
  \end{cases}
$.
\end{small}
$\mathbf{x}_{ej}^t$ is the $j$-th instance in $\mathbf{X}_e^t$, and $\text{tr}(\cdot)$ denotes the trace of a square matrix.
If $\mathbf{x}^t_{ej}$ and $\mathbf{x}^t_{ej'}$ are mutual $r$-nearest neighbours to each other, % in the original feature space,
the value of $H_{jj'}$ equals that of the kernel function $\mathcal{K}(\cdot,\cdot)$.
By maximizing the unlabeled discriminant criterion $\tau_e$, the local scatter covariance matrix guarantees the first principle, while \begin{small}$\mathbf{S}_e^N=\sum_{j,j'=1}^{n^t_e}\frac{\mathcal{K}(\mathbf{x}^t_{ej},\mathbf{x}^t_{ej'})-H_{jj'}}{(n^t_e)^2}(\mathbf{x}^t_{ej}-\mathbf{x}^t_{ej'})(\mathbf{x}^t_{ej}-\mathbf{x}^t_{ej'})^T$\end{small}, the non-local scatter covariance matrix, enforces the second principle.
%The value of
$\tau_e$ also depends on the kernel used, since different kernels indicate different neighbour %hood
information and different degrees of similarity between neighboured instances.
%, so that the value of $\tau_e$ also depends on the kernel selected.
With $\tau_{e(k)}$ obtained from the $k$-th kernel $\mathcal{K}_k$, the unlabeled discriminant criterion %in multiple kernels
$\tau_e$ can be written as $\tau_{e} = \sum_{k=1}^{N_k}\beta_k\tau_{e(k)}=\boldsymbol{\beta}^T \boldsymbol{\tau}_e$ where $\boldsymbol{\tau}_e = [\tau_{e(1)}, \cdots, \tau_{e(N_k)}]$.

\paragraph{The optimization problem}
Combining the two aspects abovementioned to model the reflection function $f$, we finally formulate the optimization problem  as follows,
\begin{small}
\begin{align}
\boldsymbol{\beta}^*, \lambda^*, \mu^*, b^*=\arg\min_{\boldsymbol{\beta}, \lambda, \mu, b}&\ \sum_{e=1}^{N_e}\mathcal{L}_h\left(\boldsymbol{\beta}^T\mathbf{\hat{d}}_e + \lambda\boldsymbol{\beta}^T\mathbf{\hat{Q}}_e\boldsymbol{\beta} + \frac{\mu}{\boldsymbol{\beta}^T\boldsymbol{\tau}_e}+b, \frac{1}{l_e}\right) + \gamma_1 R(\boldsymbol{\beta},\lambda,\mu,b), \nonumber\\
\text{s.t.}&\quad \beta_k\geq 0, \ \forall k\in\{1,\cdots,N_k\}, \ \lambda \geq 0, \ \mu\geq 0,
\end{align}
\end{small}\noindent
where $1/f = \boldsymbol{\beta}^T\mathbf{\hat{d}}_e + \lambda\boldsymbol{\beta}^T\mathbf{\hat{Q}}_e\boldsymbol{\beta} + \frac{\mu}{\boldsymbol{\beta}^T\boldsymbol{\tau}_e}+b$  and $\mathcal{L}_h(\cdot)$ is the Huber regression loss~\cite{Huber:Robust} constraining the value of $f$ to be as close to $1/l_e$ as possible.
$\gamma_1$ controls the %degree of 
complexity of the parameters. % by l2-regularization.
Minimizing the difference between domains,
%a source and a target domain,
including the MMD %distance
$\boldsymbol{\beta}^T\mathbf{\hat{d}}_e$ and the distance variance $\boldsymbol{\beta}^T\mathbf{\hat{Q}}_e\boldsymbol{\beta}$, and meanwhile maximizing the discriminant criterion $\boldsymbol{\beta}^T\boldsymbol{\tau}_e$ in the target domain will contribute a large performance improvement ratio $l_e$ (i.e., a small $1/l_e$).
$\lambda$ and $\mu$ balance the importance of the three terms in $f$, and $b$ is the bias term.
%This optimization problem is a nonlinear constrained convex problem, and can be efficiently and effectively solved.

%The other critical issue to be addressed is the inconsistency between $\mathbf{W}_e$ for $e=\{1,\cdots,N_e\}$, which makes inferring what to transfer impossible.
%For example, the semantic meanings of the $u$ latent factors of $\mathbf{W}_1$ may significantly differ from those of $\mathbf{W}_{N_e}$.
%To address the problem, we learn a global latent factor dictionary $\mathbf{W}_G\in\mathbb{R}^{m\times N_G}$ containing $N_G$ latent factors that can reconstruct each $\mathbf{W}_e$:
%\begin{align}
%&\mathbf{W}_G ^* = \arg\min_{\mathbf{W_G},\mathbf{Z}_e}\frac{1}{N_e}\sum_{e=1}^{N_e}\Vert\mathbf{W}_e^T-\mathbf{Z}_e\mathbf{W}_{G}^T\Vert^2_2 + \gamma\vert \mathbf{Z}_e\vert, \nonumber\\
%&\text{s.t.}\quad\forall g=1,\cdots, N_G, \quad
%(\mathbf{W}_{G(g)})^T\mathbf{W}_{G(g)} \leq 1,
%\end{align}
%where $\gamma$ regularizes the sparsity of coefficients $\mathbf{Z}_e$, and $\mathbf{W}_{G(g)}$ is the $g$-th column of $\mathbf{W}_G$.

\subsection{Inferring What to Transfer}
Once the L2T agent has learned the reflection function $f(\mathcal{S},\mathcal{T},\mathbf{W};\boldsymbol{\beta}^*,\lambda^*,\mu^*,b^*)$, %and the global latent factor dictionary $\mathbf{W}_G$ from previous transfer learning experiences,
it can take advantage of the function to infer the optimal what to transfer, i.e., the latent feature factor matrix $\mathbf{W}$, for a newly arrived pair of source domain $\mathcal{S}_{N_e+1}$ and target domain $\mathcal{T}_{N_e+1}$.
The optimal %what to transfer, i.e., the
latent feature factor matrix $\mathbf{W}^*$ should %minimize 
maximize
the value of $f$. % given a  future source domain $\mathcal{S}_{N_e+1}$ and a target domain $\mathcal{T}_{N_e+1}$.
%Second, $\mathbf{W}$ can be reconstructed from the global latent factor dictionary $\mathbf{W}_G$ which is assumed to contain sufficient latent factor bases.
To this end, we optimize the following objective with regard to $\mathbf{W}$,
%\begin{align}
%&\mathcal{L}(\mathbf{W}, \mathbf{Z})=
%&\sum_{k=1}^{N_k}\beta_k tr(\mathbf{K}_{\mathbf{W}}^{(k)}\mathbf{M}) + \mu\Vert\mathbf{W}^T-\mathbf{Z}\mathbf{W}_G^T\Vert_F^2 +\lambda\vert\mathbf{Z}\vert.
%\end{align}
\begin{small}
\begin{align}
\mathbf{W}^* &= \arg\max_{\mathbf{W}}\ f(\mathcal{S}_{N_e+1},\mathcal{T}_{N_e+1},\mathbf{W};\boldsymbol{\beta}^*,\lambda^*,\mu^*,b^*)
+\gamma_2\Vert\mathbf{W}\Vert^2_F\nonumber\\
&=\arg\min_{\mathbf{W}}\ (\boldsymbol{\beta}^*)^T \mathbf{\hat{d}}_\mathbf{W} + \lambda^*(\boldsymbol{\beta}^{*})^T\mathbf{\hat{Q}}_\mathbf{W}\boldsymbol{\beta}^*+\mu^*\frac{1}{(\boldsymbol {\beta}^*)^T\mathbf{\boldsymbol{\tau}}_\mathbf{W}} + \gamma_2\Vert\mathbf{W}\Vert^2_F,\label{obj_reflection}
\end{align}
\end{small}\noindent
where $\Vert\cdot\Vert_F$ denotes the matrix Frobenius norm and $\gamma_2$ controls the complexity of $\mathbf{W}$. The first and second terms in problem (\ref{obj_reflection}) can be calculated as
{\scriptsize
\begin{align*}
(\boldsymbol{\beta}^*)^T\mathbf{\hat{d}}_\mathbf{W}  = \sum_{k=1}^{N_k}\beta_k^*\biggl[\frac{1}{(n^s)^2} \sum_{i,i'=1}^{n^s}\mathcal{K}_k(\mathbf{x}^{s}_{i}\mathbf{W}, \mathbf{x}^{s}_{i'}\mathbf{W}) +\frac{1}{(n^t)^2}\sum_{j,j'=1}^{n^t}\mathcal{K}_k(\mathbf{x}^{t}_{j}\mathbf{W},\mathbf{x}^{t}_{j'}\mathbf{W})
-\frac{2}{n^s n^t}\sum_{i,j=1}^{n^s,n^t}\mathcal{K}_k(\mathbf{x}^{s}_{i}\mathbf{W},\mathbf{x}^{t}_{j}\mathbf{W})\biggr] ,
\end{align*}
}
%=\frac{1}{n^2-1}\sum_{i=1}^{n}\sum_{i'=1}^n\sum_{k=1}^{N_k}& \biggl[\beta_k^* \biggl(h_k(\mathbf{x}^s_{i},\mathbf{x}^s_{i'},\mathbf{x}^t_{i},\mathbf{x}^t_{i'})-\frac{1}{n^2}\sum_{i=1}^n\sum_{i'=1}^nh_k(\mathbf{x}^s_{i},\mathbf{x}^s_{i'},\mathbf{x}^t_{i},\mathbf{x}^t_{i'})\biggr)\biggr]^2 \nonumber \\
{\small
\begin{align*}
(\boldsymbol{\beta}^*)^T\mathbf{\hat{Q}}_\mathbf{W}\boldsymbol{\beta}^*
=& \frac{1}{n^2-1}\sum_{i,i'=1}^{n}\sum_{k=1}^{N_k}\biggl\{\beta^*_k\biggl[\mathcal{K}_k(\mathbf{x}^{s}_{i}\mathbf{W}, \mathbf{x}^{s}_{i'}\mathbf{W}) + \mathcal{K}_k(\mathbf{x}^{t}_{i}\mathbf{W}, \mathbf{x}^{t}_{i'}\mathbf{W}) - 2\mathcal{K}_k(\mathbf{x}^{s}_{i}\mathbf{W}, \mathbf{x}^{t}_{i'}\mathbf{W}) \nonumber \\
&- \frac{1}{n^2}\sum_{i,i'=1}^{n} \biggl(\mathcal{K}_k(\mathbf{x}^{s}_{i}\mathbf{W}, \mathbf{x}^{s}_{i'}\mathbf{W}) + \mathcal{K}_k(\mathbf{x}^{t}_{i}\mathbf{W}, \mathbf{x}^{t}_{i'}\mathbf{W}) - 2\mathcal{K}_k(\mathbf{x}^{s}_{i}\mathbf{W}, \mathbf{x}^{t}_{i'}\mathbf{W})\biggr)\biggr] \biggr\}^2,
\end{align*}}\noindent
where $n=\min(n^s,n^t)$. The third term in problem (\ref{obj_reflection}) can be computed as $(\boldsymbol{\beta}^*)^T\boldsymbol{\tau}_\mathbf{W} = \sum_{k=1}^{N_k}\beta^*_k\frac{\text{tr}(\mathbf{W}^T\mathbf{S}^N_k\mathbf{W})}{\text{tr}(\mathbf{W}^T\mathbf{S}^L_k\mathbf{W})}$.
%$\mu$ and $\lambda$ are the two trade-off parameters regularizing $\mathbf{W}$ and $\mathbf{Z}$, respectively.
%$\mathcal{L}$ is non-convex with the two variables variables $\mathbf{W}$ and $\mathbf{Z}$, thus only has local optimal solutions.
%However, by fixing one of them, $\mathcal{L}$ is convex with respect to the other. \\
%In this paper, we consider multiple Gaussian kernels $\mathcal{K}_k(\mathbf{a},\mathbf{b})=\text{exp}(-\Vert \mathbf{a}-\mathbf{b}\Vert^2/\delta_k)$ with different bandwidths, i.e., varying values of $\delta_k$.
%Note that $n=\min\{n^s,n^t\}$, and we calculate an approximate $\mathbf{Q}_\mathbf{w}$ for speeding up.
We optimize the non-convex problem (\ref{obj_reflection}) w.r.t %with respect to
$\mathbf{W}$ by
%We alternatively optimize the two parameters by fixing one of them each time. \\
%1) Fix $\mathbf{Z}$, optimize $\mathbf{W}$. To effectively address this, we
employing a conjugate gradient method in which the gradient is listed in the supplementary material due to page~limit.

\section{Experiments}
\vspace{-0.1in}
%\subsection{Experimental Settings}
In this section, we empirically test the performance of the proposed L2T framework.
\paragraph{Datasets} We evaluate the L2T framework on two image datasets, Caltech-256~\cite{Griffin:Caltech} and Sketches~\cite{Eitz:Hdhso}.
Caltech-256,
%originally
collected from Google Images, contains a total of 30,607 images in 256 categories.
The Sketches dataset, however, consists of 20,000 unique sketches by human beings
that
%.
%These sketches
are evenly distributed over 250 different categories.
There is an overlap between the 256 categories in Caltech-256 and the 250 categories in Sketches, e.g., ``backpack'' and ``horse''.
Obviously, the  Caltech-256 images lie in a different distribution from the Sketches%images
, so that we can regard one of them as the source domain and the other as the target domain.
%In the L2T framework, we have to collect many pairs of source and target domains, some of which are used to generate transfer learning experiences for training and some of which are required to validate and test the performance of the learnt reflection function.
We construct each pair of source and target domains by randomly sampling three categories from Caltech-256 as the source domain and randomly sampling three categories from Sketches as the target domain, which we give an example in the supplementary material.
%as shown in Figure~\ref{fig:dataset}.
%\begin{figure}[h]
%\centering
%\includegraphics[scale = 0.4]{fig/dataset}
%\caption{One example pair of source and target domains.}
%\label{fig:dataset}
%\end{figure}
Consequently, there are $20,000/250\times 3=720$ examples in the %each 
target domain
of each pair.
In total, we generate 1,000 training pairs %as
for
 transfer learning experiences, 500 validation pairs to determine hyperparameters of the reflection function, and 500 testing pairs to evaluate the reflection function.
We characterize each image from both datasets using 4,096-dimensional features extracted by a convolutional neural network pre-trained by the ImageNet dataset.

\paragraph{Baselines and Evaluation Metrics}
We compare the proposed L2T framework with the following nine baseline algorithms, including one without transfer learning and the other eight %as
feature-based transfer learning algorithms
in two categories, i.e.,
common latent space %based %and
and
manifold ensemble.
The \textbf{Original} builds a model using labeled data in the target domain only. % without transferring knowledge from the source domain.
The common latent space based algorithms are listed as follows.
The \textbf{TCA}~\cite{Pan:TCA} learns %some
transferable components across domains on which the embeddings of two domains have the minimum MMD.
The \textbf{ITL}~\cite{Shi:Information} learns a common subspace where instances from both domains are distributed not only similarly but also discriminatively via optimizing an information-theoretic metric.
The graph co-regularized matrix factorization proposed in~\cite{Long:Graph},  denoted as \textbf{CMF}, extracts a common subspace in which geometric structures within each domain and across domains are preserved.
The \textbf{LSDT}~\cite{Zhang:LSDT} learns a common subspace where instances in a target domain can be sparsely reconstructed by the combined source and target instances.
The self-taught learning (\textbf{STL})~\cite{Raina:Self} learns a dictionary from a source domain and obtains enriched representations of target instances
%in a target domain
based on the learned dictionary.
The \textbf{DIP}~\cite{Baktashmotlagh:Unsupervised} and \textbf{SIE}~\citep{Baktashmotlagh:Domain} learn a domain-invariant projection which can minimize the MMD and the Hellinger distance between domains, respectively.
%\textbf{SIE}: SIE~\citep{Baktashmotlagh:Domain} improves DIP by replacing the MMD distance with the Hellinger distance which is believed to better measure the difference between two domains.
The \textbf{GFK}~\cite{Gong:Geodesic}, a manifold ensemble based algorithm,
%: The geodesic flow kernel (GFK)
embeds a source and a target domain onto a Grassmann manifold and projections in an infinite number of subspaces along the manifold %between the two domains
are integrated to represent the target domain. The eight feature-based transfer learning algorithms %,
%in two categories, i.e.,
%either common latent space %based %and
%or
%manifold ensemble based,
are also the base transfer learning algorithms used to generate %transfer learning
experiences in Section~\ref{sec:para}.
Based on feature representations obtained by %the following
different algorithms, we use the nearest-neighbor classifier %with $k$ equal to onein order to
to perform three-class classification for the target domain.

We use the classification accuracy on the test data of the target domain as an evaluation metric.
Since target domains are at different levels of difficulty, %making
accuracies are incomparable for different pairs of domains. To evaluate the L2T on different pairs of source and target domains,
%To compare the performance over different target domains,
we adopt the performance improvement ratio defined in Section~\ref{sec:framework} as another evaluation metric.
%The performance improvement ratio is defined as $l=\frac{acc_t}{acc_o}$, where $acc_t$ is the accuracy achieved by a transfer learning algorithm while $acc_o$ is the accuracy without transfer.

%\begin{figure}[b]
\begin{wrapfigure}{r}{0.6\textwidth}
\vspace{-0.2in}
\centering
\includegraphics[scale = 0.28]{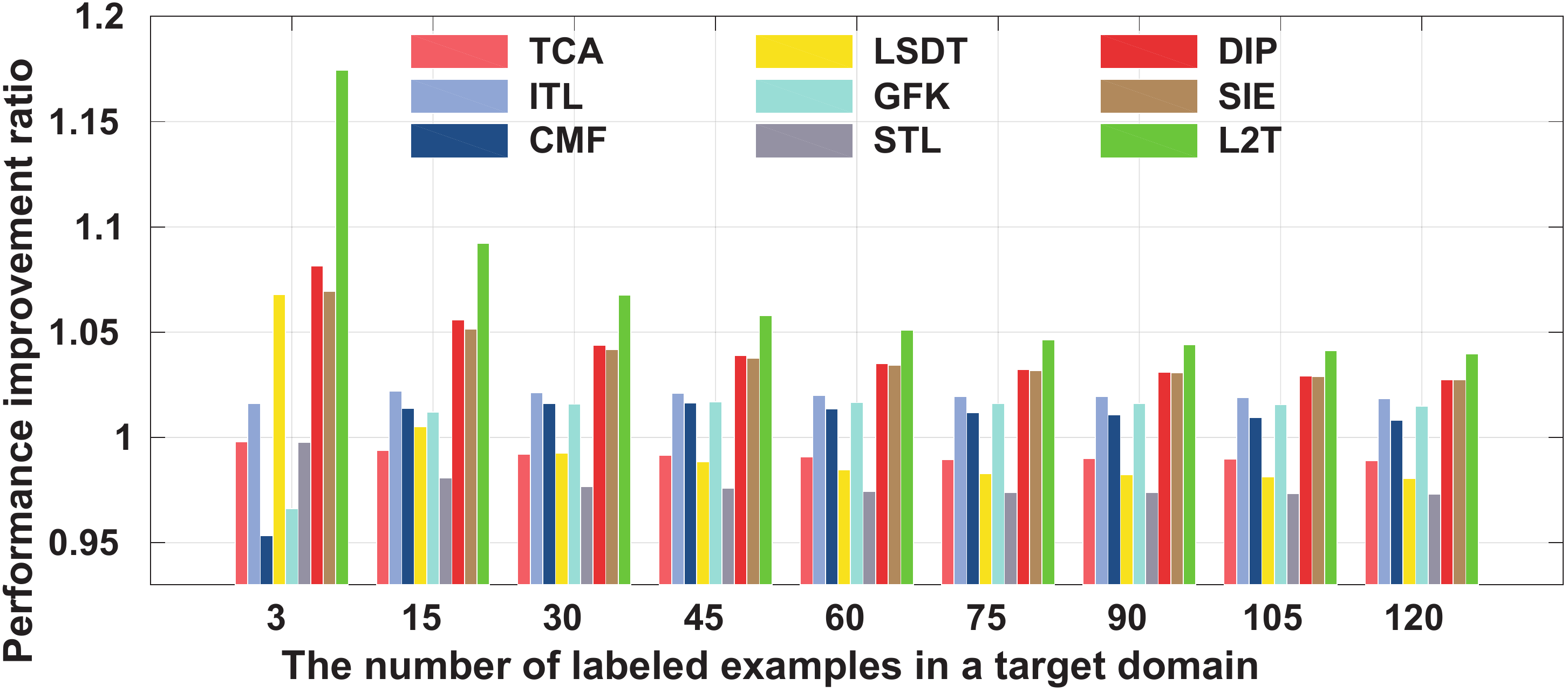}
\caption{\small{Average performance improvement ratio comparison over 500 testing pairs of source and target domains.}}
\label{fig:500_avg}
%\end{figure}
\vspace{-0.2in}
\end{wrapfigure}

%\subsection{Results}
\paragraph{Performance Comparison}
In this experiment, we learn a reflection function from 1,000 transfer learning experiences, and evaluate the reflection function on 500 testing pairs of source and target domains by comparing the 
average 
performance improvement ratio with the baselines.
In building the reflection function,  we use $33$ RBF kernels with the bandwidth $\delta_k$ in the range of $[2^{-8}\eta:2^{0.5}\eta:2^{8}\eta]$ where $\eta=\frac{1}{n^s n^t}\sum_{i,j=1}^{n^s,n^t}\Vert\mathbf{x}^s_i\mathbf{W} -\mathbf{x}^t_j\mathbf{W}\Vert^2_2 $ following the median trick~\cite{Gretton:Kernel}.
As Figure~\ref{fig:500_avg} shows, on average the proposed L2T framework outperforms the baselines up to  $10\%$ when varying the number of labeled samples in the target domain.
As the number of labeled examples in a target domain used for %training
building the classifier 
 increases from 3 to 120, the performance improvement ratio becomes smaller because the classification accuracy without transfer tends to increase.
The baseline algorithms behave differently.
%on average
The transferable knowledge learned by LSDT helps a target domain a lot when training examples are scarce,
while %the manifold ensemble method
GFK performs poorly until training examples
%are becoming more.
become more.
STL is almost the worst baseline because
%it does not learn transferable knowledge
%from a source and a target domain
%between two domains
%simultaneously.
%Instead,
it learns a dictionary from the source domain only but ignores the target domain.
%learns a dictionary from a source domain and then apply the dictionary to a target,
%which
It runs at a high risk of failure especially when two domains are distant.
%The two algorithms
DIP and SIE, which minimize the MMD and Hellinger distance between domains subject to manifold constraints, are competent.
Note that we have run the paired $t$-test between %our algorithm
L2T and each baseline with all the $p$-values %are
on the order of $10^{-12}$, concluding that the L2T is significantly superior.
\begin{figure}[t]
\captionsetup[subfigure]{justification=centering}
\centering
\begin{subfigure}{0.32\textwidth}
\includegraphics[scale=0.235]{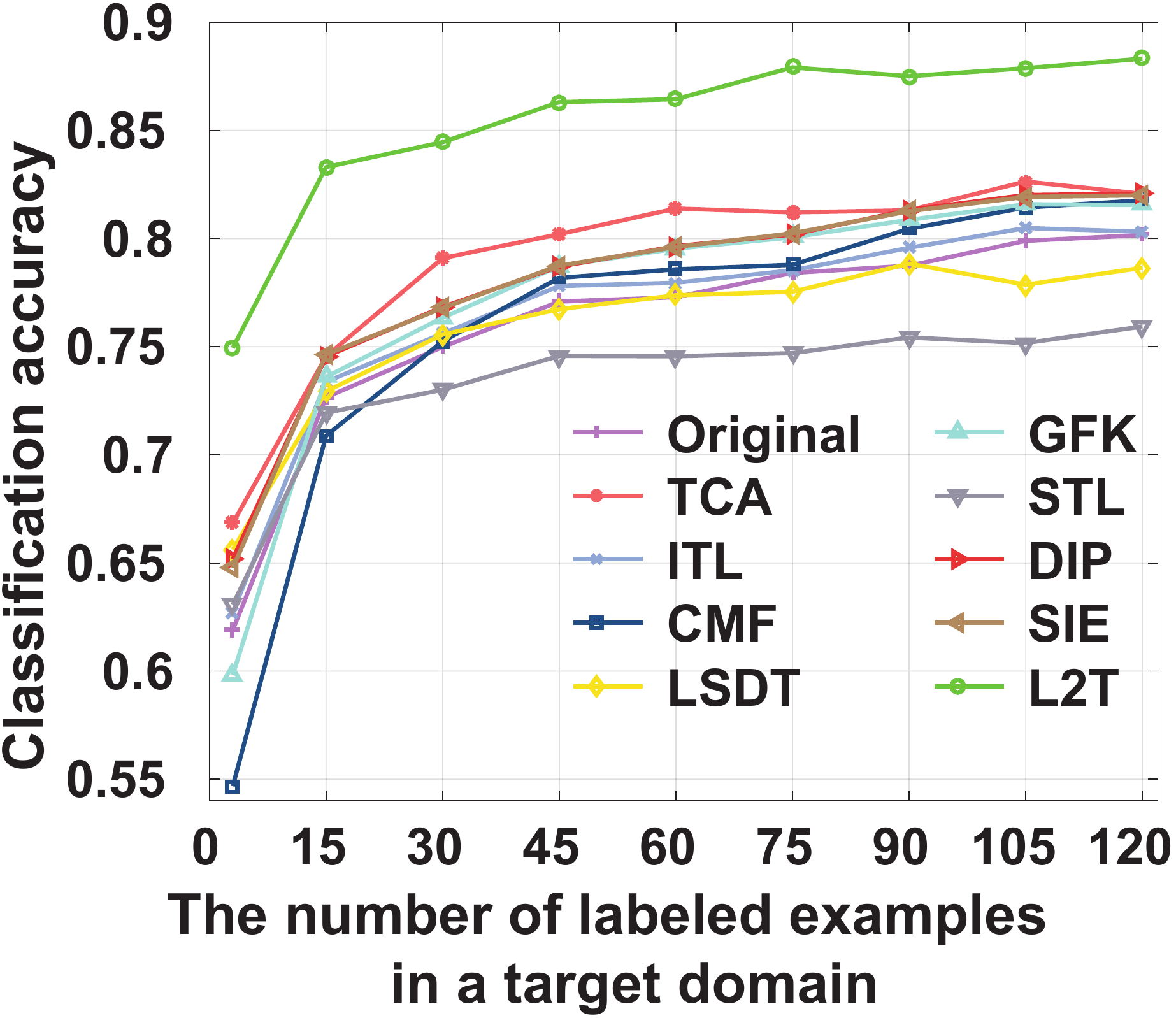}
\caption{galaxy / harpsichord / saturn \\ $\rightarrow$ kangaroo / standing-bird / sun}
\label{fig:selected_1}
\end{subfigure}
\begin{subfigure}{0.32\textwidth}
\includegraphics[scale=0.235]{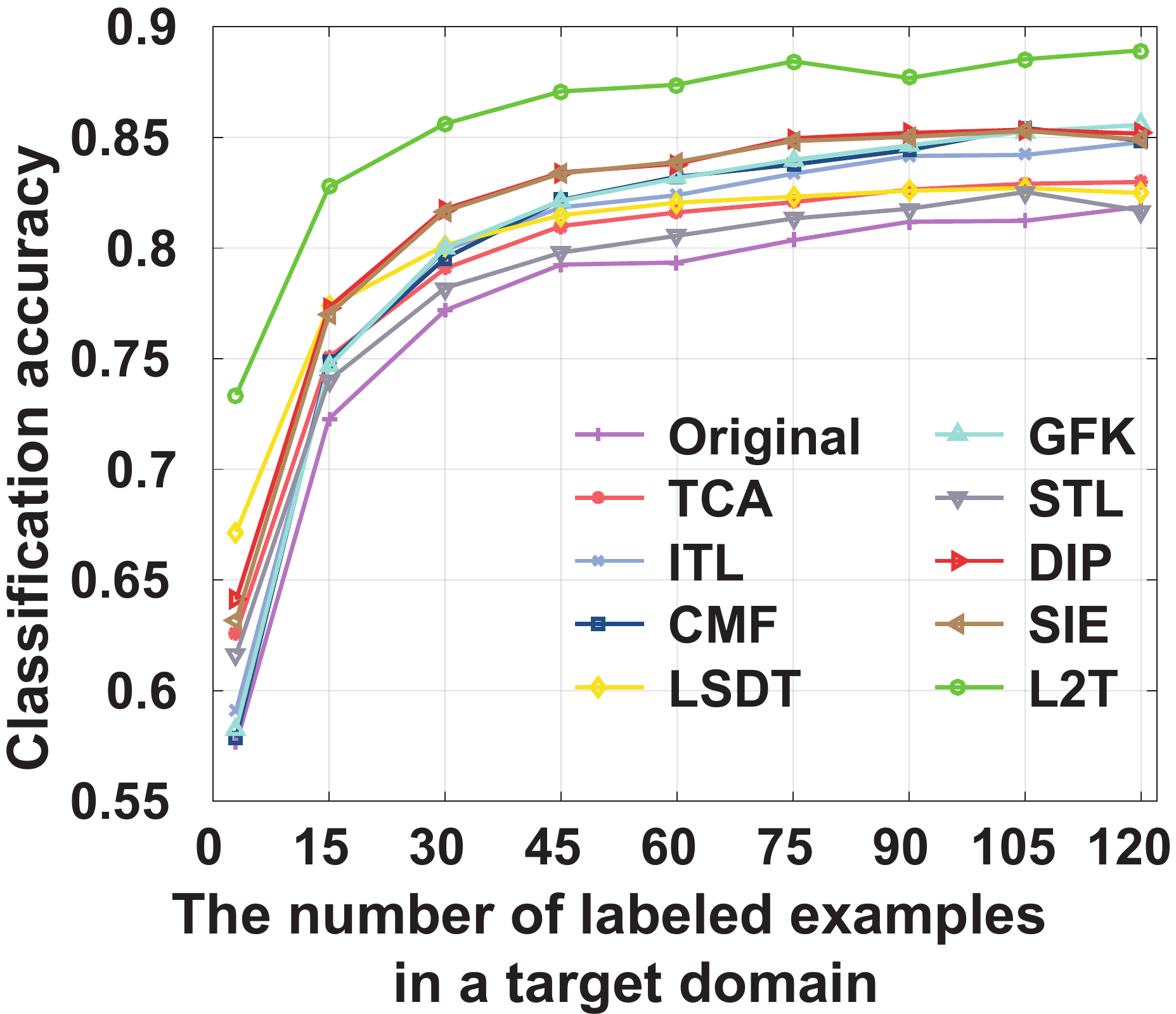}
\caption{bat / mountain-bike / saddle \\ $\rightarrow$ bush / person / walkie-talkie}
\label{fig:selected_2}
\end{subfigure}
\begin{subfigure}{0.32\textwidth}
\includegraphics[scale=0.235]{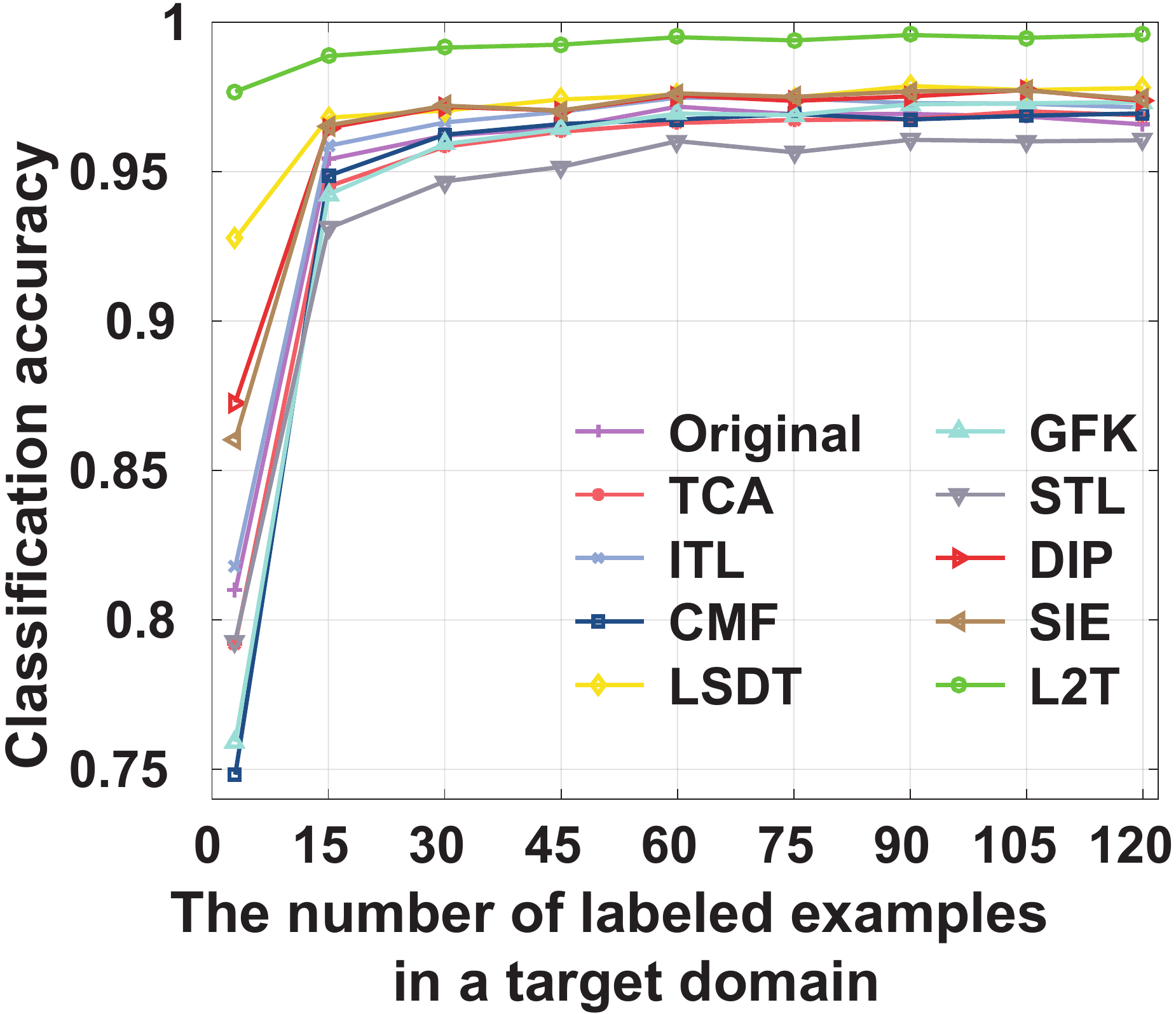}
\caption{microwave / spider / watch \\ $\rightarrow$ spoon / trumpet / wheel}
\label{fig:selected_3}
\end{subfigure}
\begin{subfigure}{0.32\textwidth}
\includegraphics[scale=0.235]{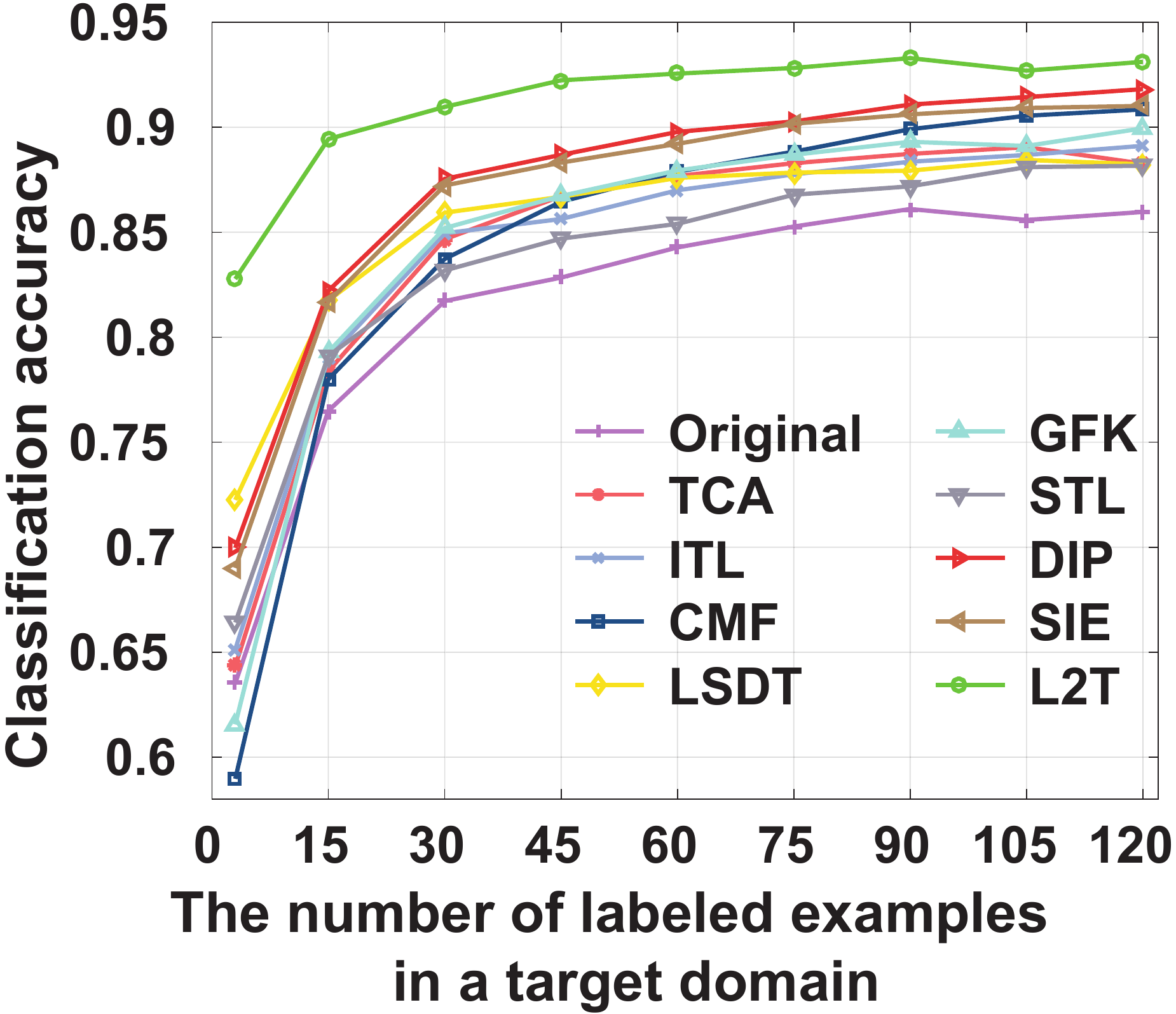}
\caption{bridge / harp / traffic-light \\ $\rightarrow$ door-handle / hand / present}
\label{fig:selected_4}
\end{subfigure}
\begin{subfigure}{0.32\textwidth}
\includegraphics[scale=0.235]{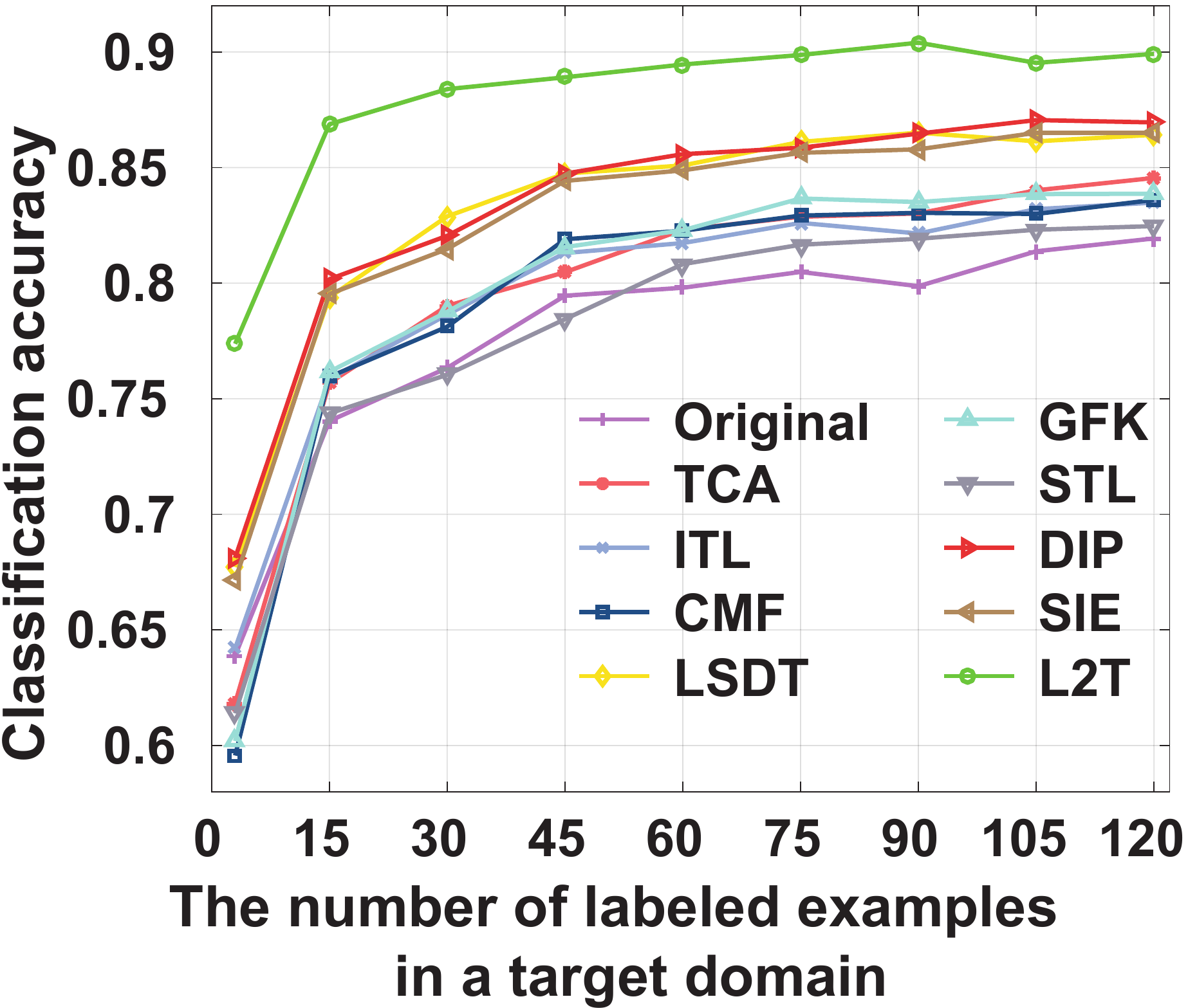}
\caption{bridge / helicopter / tripod \\$\rightarrow$ key / parrot / traffic-light}
\label{fig:selected_5}
\end{subfigure}
\begin{subfigure}{0.32\textwidth}
\includegraphics[scale=0.235]{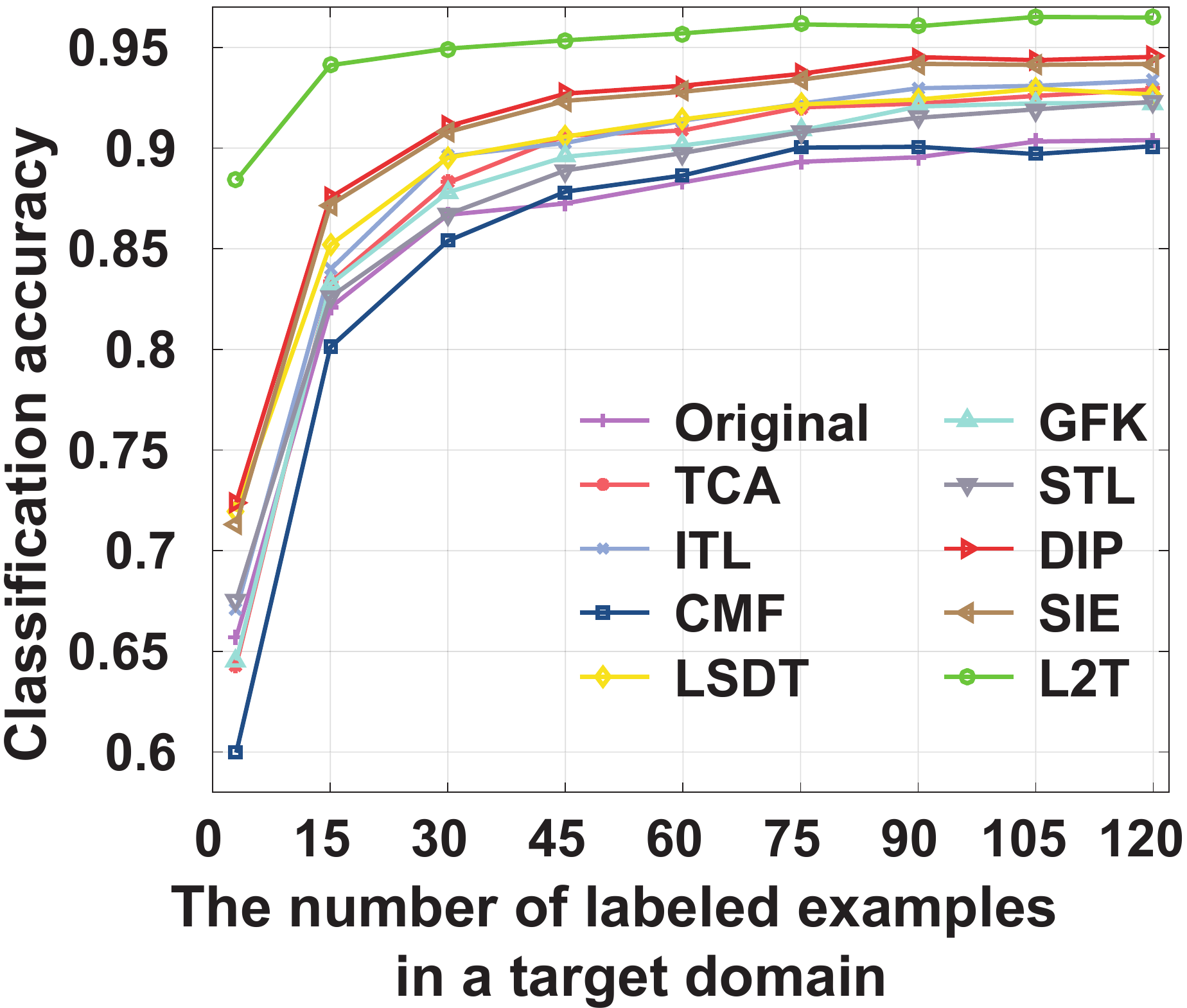}
\caption{caculator / straw / french-horn \\$\rightarrow$ doorknob / palm-tree / scissors}
\label{fig:selected_6}
\end{subfigure}
\caption{Classification accuracies on six pairs of source and target domains.}
\label{fig:selected}
\end{figure}

%For different pairs of source and target domains, %either close or distant,
%L2T improves to varying degrees.
%Therefore,
We also randomly select six of the 500 testing pairs and compare classification accuracies %of 
by different algorithms for each pair in Figure~\ref{fig:selected}.
The performance of all baselines %on the six pairs
varies from pair to pair. Among all the baseline methods, TCA performs the best when transferring between domains in Figure~\ref{fig:selected_1} and LSDT is the most superior in Figure~\ref{fig:selected_3}.
However, L2T consistently
%improves the classification accuracies in target domains
outperforms the baselines on all the settings.
For some pairs, e.g., Figures~\ref{fig:selected_1}, \ref{fig:selected_3} and \ref{fig:selected_6}, the three classes in the target domains are comparably easy to tell apart, hence the Original without transfer can %even
achieve
even better results than some transfer learning algorithms.
In this case,
L2T %can
still
%classify better
improves by discovering the best transferable knowledge from the source domain,
especially when the number of labeled examples is small (see Figure~\ref{fig:selected_3} and~\ref{fig:selected_6}).
If two domains are very related, e.g.,
%the classes of ``galaxy'' and ``saturn'' in the source and the class of ``sun'' in the target domain
the source %domain
with ``galaxy'' and ``saturn'' and the target with ``sun''
in Figure~\ref{fig:selected_1}, L2T even
finds out more transferable knowledge and contributes more significant improvement.

% Table generated by Excel2LaTeX from sheet 'Varying different approaches'
\begin{table}[t]
  \centering
  \newcommand{\tabincell}[2]{\begin{tabular}{@{}#1@{}}#2\end{tabular}}%放在导言区
  \caption{The performance improvement ratios by varying different approaches used to generate transfer learning experiences. For example, ``ITL+L2T'' denotes the L2T learning from experiences generated by ITL only, and the second line of results for ``ITL+L2T'' is the $p$-value compared to ITL.}
  \begin{adjustbox}{max width=0.92\textwidth}
    \begin{tabular}{cccccccccc}
    \toprule
    \tabincell{c}{\# of labeled\\ examples} & 3     & 15    & 30    & 45    & 60    & 75    & 90    & 105   & 120 \\
    \midrule
    TCA   & 1.0181  & 1.0024  & 0.9965  & 0.9973  & 0.9941  & 0.9933  & 0.9938  & 0.9927  & 0.9928  \\
     ITL   & 1.0188  & 1.0248  & 1.0250  & 1.0254  & 1.0250  & 1.0224  & 1.0232  & 1.0224  & 1.0224  \\
    CMF   & 0.9607  & 1.0203  & 1.0224  & 1.0218  & 1.0190  & 1.0158  & 1.0144  & 1.0142  & 1.0125  \\
    LSDT  & 1.0828  & 1.0168  & 0.9988  & 0.9940  & 0.9895  & 0.9867  & 0.9854  & 0.9834  & 0.9837  \\
    GFK   & 0.9729  & 1.0180  & 1.0232  & 1.0243  & 1.0246  & 1.0219  & 1.0239  & 1.0229  & 1.0225  \\
    STL   & 0.9973  & 0.9771  & 0.9715  & 0.9713  & 0.9715  & 0.9694  & 0.9705  & 0.9693  & 0.9693  \\
    DIP   & 1.0875  & 1.0633  & 1.0518  & 1.0465  & 1.0425  & 1.0372  & 1.0365  & 1.0343  & 1.0317  \\
    SIE   & 1.0745  & 1.0579  & 1.0485  & 1.0448  & 1.0412  & 1.0359  & 1.0359  & 1.0334  & 1.0318  \\
    \midrule
    \multicolumn{1}{c}{\multirow{2}[0]{*}{ITL + L2T}} & 1.1210  & 1.0737  & 1.0577  & 1.0506  & 1.0456  & 1.0398  & 1.0394  & 1.0361  & 1.0359  \\
    \multicolumn{1}{c}{} & 0.0000  & 0.0000  & 0.0000  & 0.0000  & 0.0000  & 0.0000  & 0.0000  & 0.0002  & 0.0002  \\
    \cline{2-10}
%    \multicolumn{1}{c}{\multirow{2}[0]{*}{GFK + L2T}} & 1.1680  & 1.0956  & 1.0727  & 1.0631  & 1.0560  & 1.0495  & 1.0478  & 1.0445  & 1.0444  \bigstrut[t]\\
%    \multicolumn{1}{c}{} & 0.0000  & 0.0000  & 0.0000  & 0.0000  & 0.0000  & 0.0000  & 0.0000  & 0.0000  & 0.0000  \\
%        \cline{2-10}
    \multicolumn{1}{c}{\multirow{2}[0]{*}{DIP + L2T}} & 1.1605  & 1.0927  & 1.0718  & 1.0620  & 1.0562  & 1.0500  & 1.0483  & 1.0461  & 1.0451  \bigstrut[t]\\
    \multicolumn{1}{c}{} & 0.0000  & 0.0000  & 0.0000  & 0.0000  & 0.0000  & 0.0000  & 0.0000  & 0.0000  & 0.0000  \\
         \cline{2-10}
    \multicolumn{1}{c}{\multirow{2}[0]{*}{\tabincell{c}{(LSDT/GFK\\/SIE) + L2T}}} & 1.1660  & 1.0973  & 1.0746  & 1.0652  & 1.0573  & 1.0506  & 1.0485  & 1.0451  & 1.0429  \bigstrut[t]\\
    \multicolumn{1}{c}{} & 0.0000  & 0.0000  & 0.0000  & 0.0000  & 0.0000  & 0.0000  & 0.0000  & 0.0000  & 0.0000  \\
        \cline{2-10}
    \multicolumn{1}{c}{\multirow{2}[0]{*}{\tabincell{c}{(TCA/ITL/CMF/GFK\\/LSDT/SIE/) + L2T}}} & 1.1712	& 1.0954 &	1.0707 & 1.0607	& 1.0529	& 1.0469 & 	1.0449 &	1.0421	& 1.0416  \bigstrut[t]\\
        \multicolumn{1}{c}{} & 0.0000 &	0.0000	& 0.0001 &	0.0001	& 0.0106 & 0.0019 & 0.0002	& 0.0047 & 0.0106 \bigstrut[t]\\
    \midrule
    all + L2T & 1.1872 &	1.1054 &	 1.0795	& 1.0699 &	 1.0616 & 1.0551 & 	1.0531	& 1.0500 &	1.0502  \\
    \toprule
    \end{tabular}%
    %\vspace{-0.45in}
    \end{adjustbox}
    %\vspace{-0.35in}
  \label{tab:exp_app}%
\end{table}%

\begin{figure}[!ht]
\vspace{-0.15in}
  \begin{minipage}[t]{0.315\textwidth}
    \centering
    \includegraphics[scale=0.66]{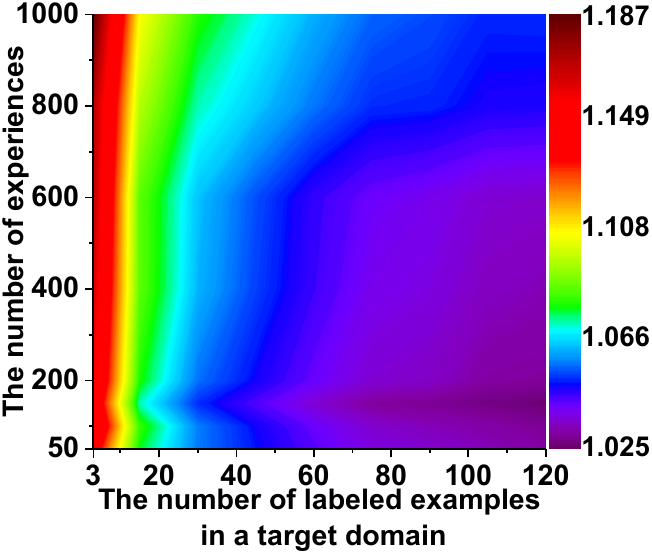}
    \figcaption{Varying the number of transfer learning experiences. }% on the performance of L2T.}
	\label{fig:exp_num}
  \end{minipage}%
  \hspace{0.1in}
  \begin{minipage}[t]{0.31\textwidth}
    \centering
    \includegraphics[scale=0.72]{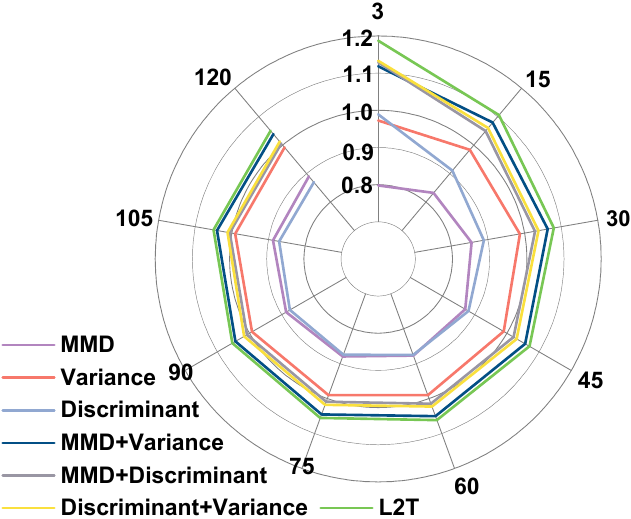}
    \figcaption{Varying the components constituted in the $f$.} %the reflection function.}
	\label{fig:exp_ref}
  \end{minipage}%
  \hspace{0.1in}
    \begin{minipage}[t]{0.31\textwidth}
    \centering
    \includegraphics[scale=0.66]{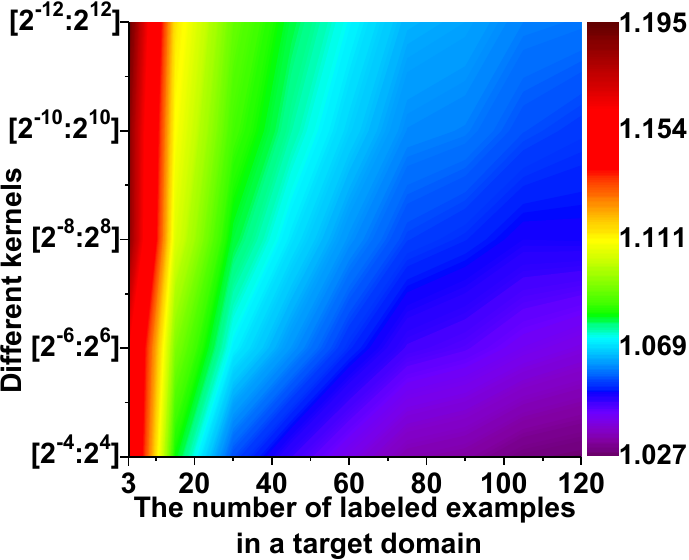}
    \figcaption{Varying the number of kernels considered in the $f$. } %reflection function.}
	\label{fig:exp_ker}
  \end{minipage}%
  \vspace{-0.15in}
\end{figure}

\paragraph{Varying the Experiences} We further investigate how transfer learning experiences used to learn the reflection function influence the performance of L2T.
In this experiment,
%Here
we evaluate on %only
50 randomly sampled %testing
pairs out of the 500 testing pairs in order to efficiently investigate a wide range of cases in the following. 
The sampled set is unbiased and sufficient to characterize such influence, evidenced by the asymptotic consistency of the average performance improvement ratio on the 500 pairs in Figure~\ref{fig:500_avg} and the 50 pairs in the last line of Table~\ref{tab:exp_app}. 
First, we fix the number of transfer learning experiences to be 1,000 and vary the set of base transfer learning algorithms. % used.
The results are shown in Table~\ref{tab:exp_app}.
Even with experiences generated by single base algorithm, e.g., ITL or DIP, the L2T can still learn a reflection function that significantly better ($p$-value < 0.05) decides what to transfer than using ITL or DIP directly.
With more base algorithms involved, the transfer learning experiences are more informative
to cover more situations of source-target pairs and the knowledge transferred between them.
As a result, the L2T learns a better reflection function
%taking more situations of a source-target pair and the knowledge transferred between them into consideration,
and thereby achieves higher performance
 improvement ratios.
Second, we fix the set of base algorithms to include all the eight baselines and vary the number of transfer learning experiences used for training.
As shown in Figure~\ref{fig:exp_num}, the average performance 
improvement 
ratio achieved by L2T tends to increase as the number of labeled examples in the target domain decreases, and more importantly it increases as the number of %transfer learning
experiences increases. This lays the foundation for further conducting L2T in an online manner which can gradually assimilate transfer learning experiences and continuously improve.

\paragraph{Varying the Reflection Function}
%Given a set of transfer learning experiences,
We also study the influence of different configurations of the reflection function on the performance of L2T.
First, we vary the components to be considered in building the reflection function $f$ as shown in Figure~\ref{fig:exp_ref}.
Considering single type, either MMD, variance, or the discriminant criterion, %incurs
brings inferior performance and even negative transfer.
L2T %which takes
taking all the three factors into consideration %simultaneously
outperforms the others, demonstrating that the three components are all necessary and mutually reinforcing.
With all the three components included, 
%L2T learns the optimal $\gamma^*=36.7564$ and $\mu^*=0.3124$. W
we plot values of the learned $\boldsymbol{\beta}^*$ in the supplementary material.
Second, we change different kernels used.
%In previous experiments, we use $33$ Gaussian kernels with the bandwidth $\delta_k$ in the range of $[2^{-8}:2^{0.5}:2^{8}]$.
In Figure~\ref{fig:exp_ker}, we present results by either narrowing down or extending the range $[2^{-8}\eta:2^{0.5}\eta:2^{8}\eta]$.
Obviously, more kernels (e.g., $[2^{-12}\eta:2^{0.5}\eta:2^{12}\eta]$),
capable of encrypting better transfer learning skills in the reflection function,
%correspond to
achieve larger performance improvement ratios.
%, indicating that better transfer learning skills are encrypted in the reflection function.

\section{Conclusion}
\vspace{-0.1in}
In this paper, we propose a novel L2T framework for transfer learning which automatically determines what and how to transfer is the best
%for a pair of
between a source and
a target domain by leveraging previous transfer learning experiences.
In particular, L2T learns a reflection function mapping a pair of domains and the knowledge transferred between them to the performance improvement ratio.
When a new pair of domains arrives, L2T optimizes what and how to transfer by maximizing the value of the learned reflection function.
We believe that L2T opens a new door to improve transfer learning by leveraging transfer learning experiences.
Many research issues, e.g., incorporating hierarchical latent feature factors as what to transfer and designing online L2T, can be further investigated.
%
%In the future, we plan to implement on online version of L2T which can gradually assimilate transfer learning experiences and continuously improve.

%\small
\bibliography{nips2017}
\bibliographystyle{abbrvnat}
\newpage
\begingroup
\setlength{\abovecaptionskip}{2pt plus 3pt minus 2pt}
\setlength{\textfloatsep}{6.5pt plus 1.0pt minus 2.0pt}

     \setlength\abovedisplayskip{10pt}%%one space
     \setlength\belowdisplayskip{10pt}%% one more
     \setlength\abovedisplayshortskip{10pt}%% a third
     \setlength\belowdisplayshortskip{10pt}%% a fourth

\vskip 0.1in
\hrule height 4pt
\vskip 0.25in
\vskip -\parskip
\begin{centering}\begin{bf}\begin{LARGE}Learning to Transfer: Supplementary Material\end{LARGE}\end{bf}\par\end{centering}
\vskip 0.29in
\vskip -\parskip
\hrule height 1pt
\vskip 0.09in%
\vskip 0.3in

%\author{
%David S.~Hippocampus\thanks{ Use footnote for providing further information
%about author (webpage, alternative address)---\emph{not} for acknowledging
%funding agencies.} \\
%Department of Computer Science\\
%Cranberry-Lemon University\\
%Pittsburgh, PA 15213 \\
%\texttt{hippo@cs.cranberry-lemon.edu} \\
%\And
%Coauthor \\
%Affiliation \\
%Address \\
%\texttt{email} \\
%\AND
%Coauthor \\
%Affiliation \\
%Address \\
%\texttt{email} \\
%\And
%Coauthor \\
%Affiliation \\
%Address \\
%\texttt{email} \\
%\And
%Coauthor \\
%Affiliation \\
%Address \\
%\texttt{email} \\
%(if needed)\\
%}
%
%% The \author macro works with any number of authors. There are two commands
%% used to separate the names and addresses of multiple authors: \And and \AND.
%%
%% Using \And between authors leaves it to \LaTeX{} to determine where to break
%% the lines. Using \AND forces a linebreak at that point. So, if \LaTeX{}
%% puts 3 of 4 authors names on the first line, and the last on the second
%% line, try using \AND instead of \And before the third author name.
%
%\newcommand{\fix}{\marginpar{FIX}}
%\newcommand{\new}{\marginpar{NEW}}
%
%%\nipsfinalcopy % Uncomment for camera-ready version
%
%\begin{document}

%\maketitle

\setcounter{section}{0}

\section{Empirical Estimation of $\mathbf{Q}_e$}
\begin{align}
\mathbf{\hat{Q}}_e
=& \frac{1}{n^2-1}\sum_{i,i'=1}^{n}\sum_{k=1}^{N_k}\biggl[\mathcal{K}_k(\mathbf{x}^{s}_{i}\mathbf{W}, \mathbf{x}^{s}_{i'}\mathbf{W}) + \mathcal{K}_k(\mathbf{x}^{t}_{i}\mathbf{W}, \mathbf{x}^{t}_{i'}\mathbf{W}) - 2\mathcal{K}_k(\mathbf{x}^{s}_{i}\mathbf{W}, \mathbf{x}^{t}_{i'}\mathbf{W}) \nonumber \\
&- \frac{1}{n^2}\sum_{i,i'=1}^{n} \biggl(\mathcal{K}_k(\mathbf{x}^{s}_{i}\mathbf{W}, \mathbf{x}^{s}_{i'}\mathbf{W}) + \mathcal{K}_k(\mathbf{x}^{t}_{i}\mathbf{W}, \mathbf{x}^{t}_{i'}\mathbf{W}) - 2\mathcal{K}_k(\mathbf{x}^{s}_{i}\mathbf{W}, \mathbf{x}^{t}_{i'}\mathbf{W})\biggr)\biggr]^2,
\end{align}
where $n=\min(n^s_e, n^t_e)$.
\section{The Gradient towards Optimizing $\mathbf{W}$}
\begin{align}
\frac{\partial\mathcal{L}}{\partial\mathbf{W}}= \frac{\partial(\boldsymbol{\beta}^*)^T\mathbf{\hat{d}}_\mathbf{W}}{\partial\mathbf{W}} + \lambda^* \frac{\partial(\boldsymbol{\beta}^*)^T\mathbf{\hat{Q}}_\mathbf{W}\boldsymbol{\beta}^*}{\partial\mathbf{W}} - \mu^* \frac{1}{[(\boldsymbol{\beta}^*)^T\boldsymbol{\tau}_\mathbf{W}]^2}\frac{\partial(\boldsymbol{\beta}^*)^T\boldsymbol{\tau}_\mathbf{W}}{\partial\mathbf{W}}
+ 2\gamma_2\mathbf{W},
\end{align}
among which
\begin{align}
\frac{\partial(\boldsymbol{\beta}^*)^T\mathbf{\hat{d}}_\mathbf{W}}{\partial\mathbf{W}} = \sum_{k=1}^{N_k}\beta_k^*\bigg[\frac{1}{(n^s)^2}\sum_{i,i'=1}^{n_s}\hat{\mathbf{K}}_{k(i,i')}^{ss}+\frac{1}{(n^t)^2}\sum_{j,j'=1}^{n_t}\hat{\mathbf{K}}_{k(j,j')}^{tt}
-\frac{2}{n^s n^t}\sum_{i,j=1}^{n_s,n_t}\hat{\mathbf{K}}_{k(i,j)}^{st})\bigg],
\end{align}
\begin{align}
\frac{\partial(\boldsymbol{\beta}^*)^T\mathbf{\hat{Q}}_\mathbf{W}\boldsymbol{\beta}^*}{\partial\mathbf{W}}
&=\frac{1}{n^2-1}\sum_{k=1}^{N_k}\sum_{i,i'=1}^{n} \hat{\mathbf{B}}_{k(i,i')} \bigg[\hat{\mathbf{K}}_{k(i,i')}^{ss}+\hat{\mathbf{K}}_{k(i,i')}^{tt}
-2\hat{\mathbf{K}}_{k(i,i')}^{st}) \nonumber \\
&- \sum_{i,i'=1}^{n} \bigg(\hat{\mathbf{K}}_{k(i,i')}^{ss}+\hat{\mathbf{K}}_{k(i,i')}^{tt}
-2\hat{\mathbf{K}}_{k(i,i')}^{st})\bigg)\bigg]
\label{eqn:dQ}
\end{align}
\begin{align}
\frac{\partial(\boldsymbol{\beta}^*)^T\boldsymbol{\tau}_\mathbf{W}}{\partial\mathbf{W}} = \sum_{k=1}^{N_k}\beta_k\frac{2[\text{tr}(\mathbf{W}^T\mathbf{S}^L_k\mathbf{W})]\mathbf{S}^N_k-2[\text{tr}(\mathbf{W}^T\mathbf{S}^N_k\mathbf{W})]\mathbf{S}^L_k}{[\text{tr}(\mathbf{W}^T\mathbf{S}^L_k\mathbf{W})]^2},
\end{align}
where $\hat{\mathbf{K}}^{ss}_{k(i,i')}$, $\hat{\mathbf{K}}^{tt}_{k(j,j')}$, and $\hat{\mathbf{K}}^{st}_{k(i,j)}$ depending on the kernel function are calculated as follows,
%In this paper, we consider multiple Gaussian kernels $\mathcal{K}_k(\mathbf{a},\mathbf{b})=\text{exp}(-\Vert \mathbf{a}-\mathbf{b}\Vert^2/\delta_k)$ with different bandwidths, i.e., varying values of $\delta_k$.
%As a result,
\begin{equation}
\hat{\mathbf{K}}^{ss}_{k(i,i')}=-\frac{2}{\delta_k}\mathcal{K}_k(\mathbf{x}^s_i\mathbf{W},\mathbf{x}^s_{i'}\mathbf{W})(\mathbf{x}^s_i-\mathbf{x}^s_{i'})(\mathbf{x}^s_i-\mathbf{x}^s_{i'})^T\mathbf{W}
,
\end{equation}
\begin{equation}
\hat{\mathbf{K}}^{tt}_{k(i,i')}=-\frac{2}{\delta_k}\mathcal{K}_k(\mathbf{x}^t_i\mathbf{W},\mathbf{x}^t_{i'}\mathbf{W})(\mathbf{x}^t_i-\mathbf{x}^t_{i'})(\mathbf{x}^s_i-\mathbf{x}^s_{i'})^T\mathbf{W}
,
\end{equation}
\begin{equation}
\hat{\mathbf{K}}^{st}_{k(i,i')}=-\frac{2}{\delta_k}\mathcal{K}_k(\mathbf{x}^s_i\mathbf{W},\mathbf{x}^s_{i'}\mathbf{W})(\mathbf{x}^s_i-\mathbf{x}^t_{i'})(\mathbf{x}^s_i-\mathbf{x}^t_{i'})^T\mathbf{W}
.
\end{equation}
%and $\hat{\mathbf{K}}^{tt}_{k(j,j')}$ and $\hat{\mathbf{K}}^{st}_{k(i,j)}$ can be computed similarly.
%2) Fix $\mathbf{W}$, optimize $\mathbf{Z}$. The problem $w.r.t$  $\mathbf{Z}$ now turns to a sparse coding problem as below and can be efficiently solved.
%\begin{equation}
%\mathcal{L}_{\mathbf{Z}} = \mu\Vert\mathbf{W}^T-\mathbf{Z}\mathbf{W}^T_{G}\Vert^2_F +\lambda\vert\mathbf{Z}\vert.
%\end{equation}
In Eqaution~\ref{eqn:dQ},
\begin{align}
\hat{\mathbf{B}}_{k(i,i')} &= \sum_{k=1}^{N_k}\sum_{i,i'=1}^{n} 2\beta^*_k\bigg[\mathcal{K}_k(\mathbf{x}^{s}_{i}\mathbf{W}, \mathbf{x}^{s}_{i'}\mathbf{W}) + \mathcal{K}_k(\mathbf{x}^{t}_{i}\mathbf{W}, \mathbf{x}^{t}_{i'}\mathbf{W}) - 2\mathcal{K}_k(\mathbf{x}^{s}_{i}\mathbf{W}, \mathbf{x}^{t}_{i'}\mathbf{W}) \nonumber \\
&- \frac{1}{n^2}\sum_{i,i'=1}^{n} (\mathcal{K}_k(\mathbf{x}^{s}_{i}\mathbf{W}, \mathbf{x}^{s}_{i'}\mathbf{W}) + \mathcal{K}_k(\mathbf{x}^{t}_{i}\mathbf{W}, \mathbf{x}^{t}_{i'}\mathbf{W}) - 2\mathcal{K}_k(\mathbf{x}^{s}_{i}\mathbf{W}, \mathbf{x}^{t}_{i'}\mathbf{W}))\bigg].
\end{align}
\section{Exemplar of A Pair of Source and Target Domains}
\begin{figure}[!h]
\centering
\includegraphics[scale = 0.55]{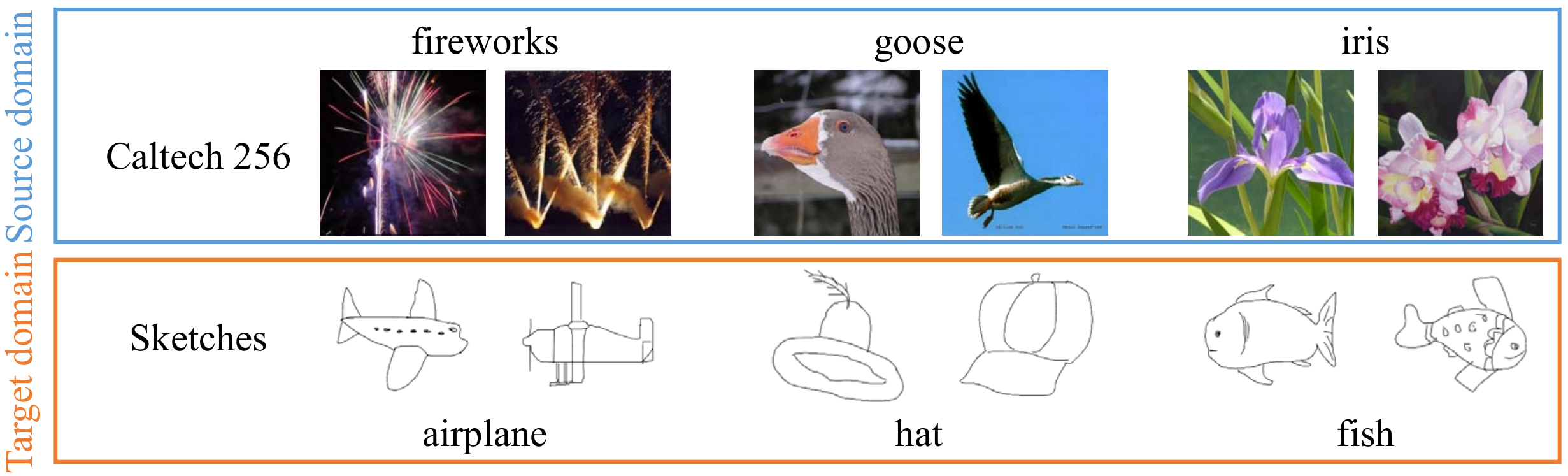}
\caption{One example pair of source and target domains.}
\label{fig:dataset}
\end{figure}
\section{Coefficients of RBF Kernels}
We plot the values of the coefficients for $N_k$ RBF kernels, i.e., $\beta_k$ for $k=\{1,\cdots,N_k\}$ in Figure~\ref{fig:beta}.
Note that we use 33 RBF kernels as stated in the paper.
\begin{figure}[h]
\centering
\includegraphics[scale = 0.4]{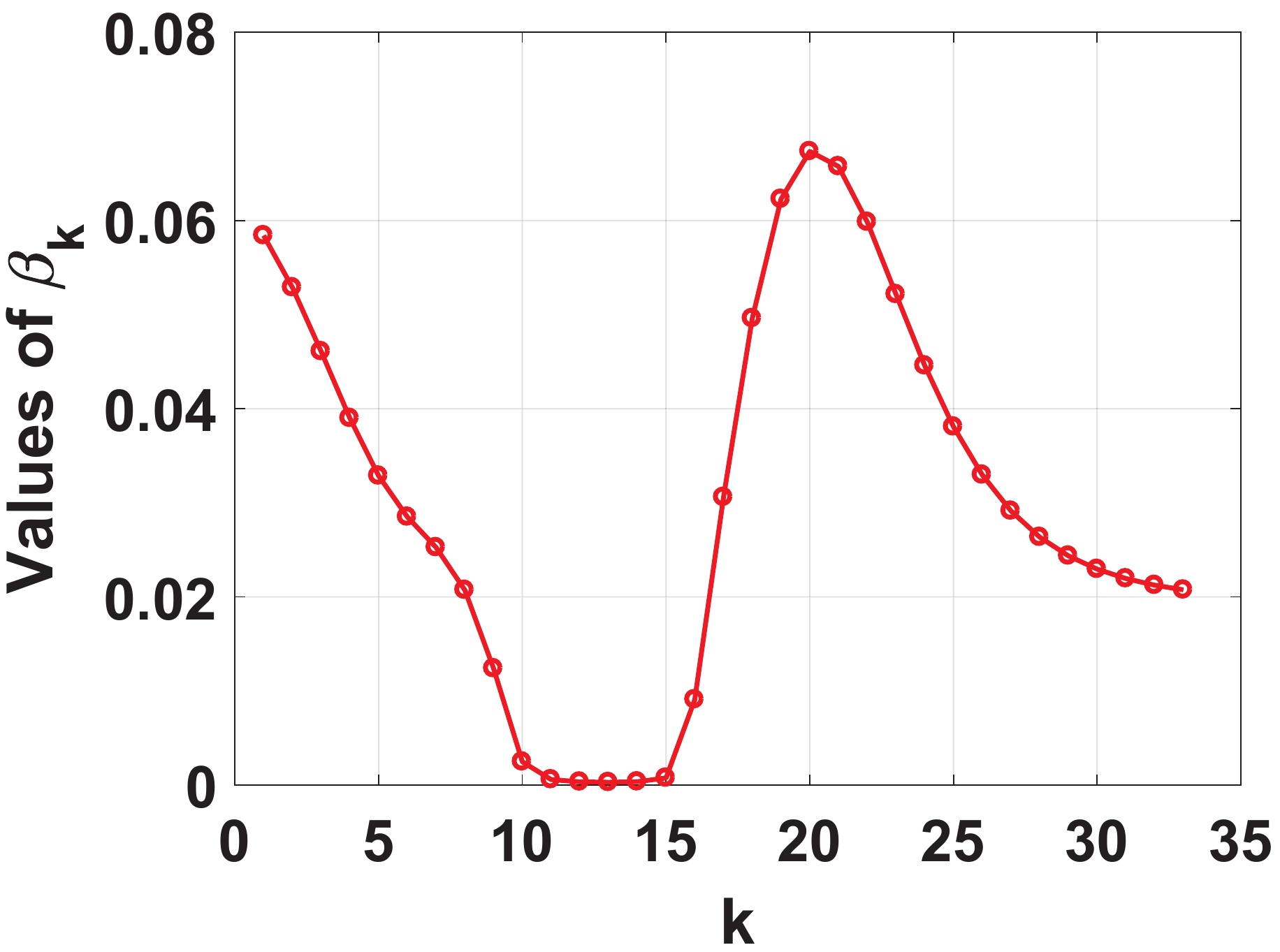}
\caption{Values of the coefficients for all $N_k$ RBF kernels.}
\label{fig:beta}
\end{figure}

\section{Discussion on $l_e$ in Equation~(2)}
$l_e$, the performance improvement ratio, heavily depends on the number of labeled examples in the target domain $\mathcal{T}_e$,  i.e., $n^t_{le}$.
A smaller number of target labeled examples tends to produce a larger performance improvement ratio, and vice versa.
Since the $n^t_{le}$ varies from experience to experience, 
we adopt a transformed $\hat{l}_e$ instead of $l_e$ to train the reflection function in Equation~(2).
The $\hat{l}_e$ is expected to be the expectation of the performance improvement ratio in the range of $[p,q]$ for the $e$-th experience, where $p$ and $q$ are the minimum and maximum number of target labeled examples.
To compute the expectation, we first assume that the performance improvement ratio with regard to the number of target labeled examples follows the following monotonically decreasing function,
\begin{equation}
f(x) = \frac{x}{a_e+b x},
\end{equation}
where  $a_e$ and $b$ are two parameters deciding the function. $a_e$ is conditioned on a specific experience but  $b$ is shared across all experiences.
Note that $f(n^t_{le})=l_e$.
As a consequence, we can obtain the expected performance improvement ratio as:
\begin{equation}
\hat{l}_e = \frac{1}{q-p}\int_p^q\frac{x}{a_e+b x} dx = \frac{1}{a_e} - \frac{b}{a_e(q-p)}\log\frac{q+b}{p+b}.
\end{equation}
Combining with the fact that $f(n^t_{le}) =\frac{n^t_{le}}{a_e+b n^t_{le}} = l_e$, we can finally obtain the corrected $\hat{l}_e$ as,
\begin{equation*}
\hat{l}_e = l_e\frac{n^t_{le}+b}{n^t_{le}}\bigg(1-\frac{b}{q-p}\log\frac{q+b}{p+b}\bigg),
\end{equation*}
where the parameter $b$ can be learned simultaneously during optimizing the objective~(2).
\endgroup

\end{document}